\documentclass[3p]{elsarticle} 

\usepackage{lineno,hyperref}

\usepackage{color,soul}

\usepackage{eurosym}

\usepackage{xcolor}

\usepackage{rotating} 

\usepackage{amsmath} 

\usepackage{booktabs}

\usepackage{colortbl}

\usepackage{amsbsy}

\usepackage{amssymb}

\usepackage{url}

\usepackage{pifont}

\usepackage{multirow}

\usepackage{textcomp} 

\modulolinenumbers[5]

\definecolor{green}{rgb}{0.1,0.4,0.1}

\journal{Journal of \LaTeX\ Templates}

\begin{document}

  
\begin{frontmatter}

\title{A lightweight and accurate YOLO-like network for small target detection in Aerial Imagery}



\author[flysight_address]{Alessandro Betti\corref{correspondingauthor}}
\ead{alessandrobetti53@yahoo.it}

\address[flysight_address]{FlySight S.r.l, Via Lampredi 45, 57121 Livorno, Italy}

\begin{abstract} 
Despite the breakthrough deep learning performances achieved for automatic object detection, small target detection is still a challenging problem, especially when looking at fast and accurate solutions suitable for mobile or edge applications.
In this work we present YOLO-S, a simple, fast and efficient network for small target detection. The architecture exploits a small feature extractor based on Darknet20, as well as skip connection, via both bypass and concatenation, and reshape-passthrough layer to alleviate the vanishing gradient problem, promote feature reuse across network and combine low-level positional information with more meaningful high-level information.
To verify the performances of YOLO-S, we build "AIRES", a novel dataset for cAr detectIon fRom hElicopter imageS acquired in Europe, and set up experiments on both AIRES and VEDAI datasets, benchmarking this architecture with four baseline detectors. Furthermore, in order to handle efficiently the issue of data insufficiency and domain gap when dealing with a transfer learning strategy, we introduce a transitional learning task over a combined dataset based on DOTAv2 and VEDAI and demonstrate that can enhance the overall accuracy with respect to more general features transferred from COCO data. 
YOLO-S is from 25\% to 50\% faster than YOLOv3 and only 15-25\% slower than Tiny-YOLOv3, outperforming also YOLOv3 in terms of accuracy in a wide range of experiments. Further simulations performed on SARD dataset demonstrate also its applicability to different scenarios such as for search and rescue operations. Besides, YOLO-S has an 87\% decrease of parameter size and almost one half FLOPs of YOLOv3, making practical the deployment for low-power industrial applications.
\end{abstract}

%


\end{frontmatter}


\section{Introduction}\label{Introduction}

Small target detection in Aerial Imagery has become a hot research topic in recent years for several applications including for example traffic monitoring, urban planning and security services~\cite{Qu_2017}\cite{Tang_2017}\cite{Ju_2019}, as well as for search and rescue (SAR) operations~\cite{Sambolek_2021}. Indeed, the recent advent of data-enabling technology such as Unmanned Aerial Vehicles (UAVs) represents a cost-effective solution for a broad customer base, satisfying a wide and almost limitless range of user needs depending on the camera axis, altitude of the craft and type of film used. Furthermore,
the growing availability of publicly available vehicles data from either satellite or UAV-carried sensors, such as VEDAI~\cite{Razakarivony_2015}, DOTA~\cite{Xia_2018} and Munich~\cite{Liu_2015} datasets, has pushed forward the research in the field.

Nevertheless, the low vehicle resolution, the poor distinctive features of tiny targets, the variability in vehicle type, size and color, as well as the presence of cluttered background or disturbing atmospheric agents such as haze or fog, represent still a challenge for satisfactory vehicle detection rate of Convolutional Neural Networks (CNNs). Moreover, occurrence of confusing objects, like containers, buildings, or road marks, may enhance the likelihood of false alarms, thus degrading the overall precision of the model.

In addition, when dealing with detection, a reasonable trade-off between accuracy and latency time is necessary. Popular object detectors are computationally demanding and memory hungry and can be executed generally only in centralized high-performance platforms.
In particular, double-stage detectors, such as R-CNN~\cite{Girshick_2014}, Fast R-CNN~\cite{Girshick_2015} and Faster R-CNN~\cite{Ren_2015}, are not suitable for Real-Time detection. 
On the other hand, single-stage detectors, such as YOLO9000~\cite{Redmon_2017}, YOLOv3~\cite{Redmon_2018}, YOLOv4~\cite{Bochkovskiy_2020} and SSD~\cite{Liu_2016}, provide real-time performances only on powerful resources.
None of them is also adequately tailored for small target detection.

From a practical point of view, many industrial applications require often the deployment of CNNs locally on edge devices close to the data source for several reasons, including for example a cheaper and faster data processing, unreliable data exchange with a remote server, or security and privacy issues. However, such devices are usually characterized by limited hardware resources in terms of performance, cost and energy, and do not include dedicated GPUs.  
 
Hence, it becomes essential the design of fast and lightweight CNN architectures, while keeping satisfactorily accuracy on small target detection.
The general-purpose and very high-speed Tiny-YOLOv3~\cite{YOLO} performs modestly on small vehicles detection due to the poor features extracted by its backbone and the coarseness of its output scales 13$\times$13 and 26$\times$26. 

Alternative solutions are now slowly emerging in Literature to detect small vehicles more precisely and quickly. 
He et al~\cite{He_2019} propose an improved version of Tiny-YOLOv3, namely TF-YOLO, which keeps the same backbone but introduces one more output scale 52$\times$52, as in YOLOv3, and lateral connection among multiple layers to improve target position. By also estimating more robust anchors by means of k-means clustering based on Jake's distance, on NWPU VHR-10 dataset~\cite{Cheng_2014}\cite{Cheng_2016} TF-YOLO outperforms Tiny-YOLOv3 with a mean Average Precision (mAP) almost 6\% higher and a similar speed of roughly 24 Frames Per Second (FPS). Moreover, on a subset of VOC2007 dataset~\cite{Everingham_2010}, TF-YOLO obtains a mAP of 31.5\% and a speed of 11.1 FPS, resulting 24.4\% less accurate and almost 30.8\% faster than YOLOv3, and 4.3\% more accurate and 10\% slower than Tiny-YOLOv3. 

Manual design of network architectures may include not vital layers for optimal and fast detection. Zhang et al~\cite{Zhang_2019} conceive therefore an automated iterative procedure of incremental model pruning based on four main steps: (i) apply channel-wise sparsity to identify the less important channels in each convolutional layers, (ii) remove the useless channels based on predefined thresholds, (iii) then fine-tune the model, (iv) finally evaluate the pruned model to determine the suitability for deployment, otherwise restart from (i). Pruning is applied to YOLOv3-SPP3, which is a modified version of YOLOv3 with Spatial Pyramid Pooling (SPP)~\cite{He_2015} added on top of the three heads in order to extract superior multi-scale features. The lightest pruned model, namely SlimYOLOv3-SPP3-95, gets a mAP on images 416$\times$416 of the VisDrone2018-Det dataset~\cite{Zhu_2019} with a 18\% relative drop compared to YOLOv3-SPP3, but resulting 80\% faster and occupying just the 8\% (59\%) of YOLOv3-SPP3's (Tiny-YOLOv3's) volume.

Ju et al~\cite{Ju_2019} introduce a simple, fast and accurate network composed by 31 convolutional layers and one reshape-passthrough layer, and a single output scale 64$\times$64 for an input image of size 512$\times$512. To expand quickly and efficiently the receptive field and get more contextual informations around targets avoiding any information loss, they implement dilated convolution instead of downsampling based on strided convolution. Furthermore, they employ reshape - passthrough layers and feature fusion to merge features from earlier layers with those of deeper layers and improve the overall localization performances.
On VEDAI dataset the network achieves a mAP of 47.8\%, i.e. 8.0\% less accurate (30.0\% more accurate) than YOLOv3 (Tiny-YOLOv3). 
On DOTA dataset, it exhibits an accuracy as high as YOLOv3 and outperforms Tiny-YOLOv3. On an Intel 
i7-5930k processor and TITAN X GPU, it can process almost 75 FPS, resulting about 5 times quicker than YOLOv3 and nearly fast as Tiny-YOLOv3.

In Ref.~\cite{Tajar_2021} authors observe that, assuming images acquired from a camera besides the road, rather than from UAV, the background does not change significantly. Hence, in this constrained environment batch normalization (BN) layers after convolution operations are not necessary. As a consequence, starting from Tiny-YOLOv3, authors trim manually and incrementally BN layers as well as whole convolutional layers which do not precede max pooling layers. The lightest trimmed network they obtain is tested over the BIT-vehicle dataset~\cite{BIT_url}, where targets may extend up to a few tens percent of the image size, achieving a mAP very close to YOLOv3 and with a speed a bit higher than Tiny-YOLOv3. 

Other research works are instead more focused on accuracy, at the expense of a processing speed and a network size unsuitable for deployment on low-power systems. For example, in Ref.~\cite{Cao_2020}, in order to make the receptive field of YOLOv3 smaller and thus more sensitive to small targets, the authors add a fourth output scale 104$\times$104. Furthermore, they introduce a L2 regularization term in the YOLOv3 loss function to avoid overfitting on training data. On a subset of DOTA dataset composed by ships and planes, they obtain a mAP improvement up to 3\% with respect to YOLOv3, despite of a slower inference due to the larger network size. 

In Ref.~\cite{Yang_2020}, based on observation that YOLOv3 mainly detects small targets at the scale 52$\times$52, authors propose a YOLO-E network based on two outputs 52$\times$52 and 104$\times$104 and implement a two-way residual submodule in order to make the network less deep. They also improve the sensitivity to target position by replacing the Intersection over Union (IoU) metric with GIoU~\cite{Rezatofighi_2019} and adding a new term 1 - GIoU to the YOLOv3 loss function. On VEDAI dataset, YOLO-E obtains a mAP of 91.2\%, almost a fifth more accurate and 6.7\% slower than YOLOv3.
In Ref.~\cite{Zhong_2017} a cascaded detector composed by two CNNs based on the VGG16 architecture~\cite{Simonyan_2014} is presented: first a set of candidate vehicle-like regions are selected by exploiting a combination of hierarchical feature maps and then the expected vehicle targets are identified and located. Detailed results on VEDAI and Munich datasets shows that the cascaded architecture outperforms Faster R-CNN equipped with different feature extractor networks in terms of accuracy, but resulting from 20\% to 30\% slower.

Also, low-resolution aerial imagery worsens the extraction of meaningful features from vehicles due to their appearance ambiguity and similarities with the context. 
To handle this problem, Ref.~\cite{Mostofa 2020} proposes a Joint Super-Resolution and Vehicle Detection Network (Joint-SRVDNet) that leverages complementary information of the two inter-related super-resolution and detection tasks. Joint-SRVDNet is composed by two main modules: a multi-scale MsGAN for image super-resolution with a 4x upsampling factor and YOLOv3 for vehicles detection. Specifically, authors demonstrate that a joint-learning of the two networks allows to obtain more meaningful targets and a higher perceptual quality in the super-resolved image, which in turn lead to an enhanced accuracy in the detection task and performances on low-resolution aerial imagery close to the existing state-of-the-art methods fed with the corresponding high-resolution aerial images.

Hence, it is clear that fast and accurate small vehicle detection remains nowadays a debated issue that encourages further research in this area. In particular, as far as detection of tiny targets from aerial images is concerned, Tiny-YOLOv3 does not guarantee an adequate accuracy~\cite{Ju_2019}. In this work we address the problem by presenting a novel YOLO-like network, namely YOLO-S or YOLO-\emph{small}, and evaluating its performances against some well-known baseline detectors. More specifically, the contributions of this paper are the following: \\
(i) We propose YOLO-S, which is really a small, simple, fast and efficient network. It employs a small feature extractor Darknet20 and a single fine-grained output scale, as well as it makes use of residual connection and feature fusion of 4$\times$, 8$\times$ and 16$\times$ down-sampled feature maps via upsampling and reshape-passthrough layers in order to alleviate the problem of gradient disappearance, strengthen feature propagation and reuse, and improve target position accordingly. Further, it adopts a lightweight convolutional set in the head sub-network to increase speed and reduce network size; \\ 
(ii) In order to make the performance benchmarking more challenging, we designed also YOLO-L, or YOLO-\emph{large}, which is a CNN detecting at three different resolution levels corresponding to 4$\times$, 8$\times$ and 16$\times$ down-scaled layers, but more focused on accuracy and suitable only for offline data processing due to the FPS close to YOLOv3; \\
(iii) We prepared two different vehicle datasets for experiments: the publicly available VEDAI and "AIRES", a novel dataset we composed for cAr detectIon fRom hElicopter imageS and annotated with precise bounding boxes. Also, we make experiments on the recently presented SARD dataset~\cite{Sambolek_2021} to verify the YOLO-S capability to generalize to the context of SAR operations. A thoroughly	explorative analysis of their main characteristics is also discussed; \\
(iv) We extensively compare YOLO-S with four baseline detectors: the well-known State of the Art YOLOv3~\cite{Redmon_2018}, Tiny-YOLOv3 and Ref.~\cite{Ju_2019}, as well as the proposed YOLO-L. Experiments are performed by either applying a sliding window based inference or a detection on full-sized images and adopting both the PASCAL VOC~\cite{Everingham_2015} and COCO~\cite{Lin_2014} evaluation protocols to make the analysis more exhaustive. \\
(v) To handle the issue of data insufficiency for vehicle detection, we also propose a double-stage training procedure by first finetuning on a dataset composed by DOTA and VEDAI, and then by training on the specific vehicle dataset of interest. In fact, usually a domain knowledge transfer is realized by finetuning a learning model starting from pre-trained weights on COCO or ImageNet~\cite{Deng_2009} datasets. However, we demonstrate that, since large publicly available datasets do not include small vehicles from aerial images, the basic single-stage trained model may be less efficient due to a domain gap and can under-perform with respect to the proposed double-stage training. 

The comparative experiments show that the proposed YOLO-S is really a cost-effective solution for practical applications, outperforming YOLOv3, Tiny-YOLOv3 and~\cite{Ju_2019} in terms of accuracy, and resulting also up to 50\% faster than YOLOv3 and competitive with Tiny-YOLOv3 and~\cite{Ju_2019}.

The rest of the paper is organized as follows: Section 3 explores the adopted datasets, Section 4 presents the proposed networks, the baseline detectors, the implemented experiments and the adopted metrics, Section 5 discusses the results, both quantitatively and qualitatively, whereas conclusions  are summarized in Section 6.  

\section{Datasets description}
\subsection{The proposed vehicles dataset}\label{Dataset_ours}
In this paper we introduce AIRES, a new vehicle database composed by aerial FHD visible images 1920$\times$1080 streamed by a WESCAM MX-15 EO/IR imaging system placed in a multi-sensor 
4-axis gyro-stabilized turret system and mounted at the front end of a manned police helicopter AW169. The helicopter flew at different altitudes from almost 300~m to 1000~m and different camera  angles ranging from about 5\textdegree ~to 80\textdegree. 
The images were acquired from June to September 2019 in two different geographical areas: the region of "Lombardia"  in  the  northern of Italy and the city of Oslo in Norway. The dataset is composed by 1275 images containing 15247 annotated ground truth (GT) objects organized in eight classes: "Van", "Truck", "Car", "Motorbike", "Person", "Other", "Boat" and "Bus".
Images have been annotated by means of the LabelImg software~\cite{LabelImg} using YOLO convention, drawing a tight bounding box around the vehicle and labelling the box according to the corresponding category. 
The statistics is summarized in Tab.~\ref{tab:statistics_dataset}: the majority class is "Car", representing the 69.0\% of the global dataset, followed by "Van" (13.2\%), "Person" (6.5\%) and "Truck" (4.9\%). The minority classes are instead "Motorbike" (0.5\%) and "Other" (0.8\%), which includes bulldozer and other ground moving vehicles employed in construction sites.

\begin{table*}[!b]
\small
  \centering
  	\begin{tabular}{c|ccc|ccc|ccc}
    \hline
    &  & \textbf{AIRES} & & & \textbf{VEDAI} & & &  \textbf{SARD} & \\
    \hline
    \textbf{Class} & \textbf{Global} & \textbf{Train } & \textbf{Test } & \textbf{Global} & \textbf{Train } & \textbf{Test } & \textbf{Global} & \textbf{Train } & \textbf{Test } \\
    \hline
    \textbf{Van} & 2012  & 1437 & 575  & 498 & 350 &  148 & - & - & - \\
    \hline
    \textbf{Truck} & 755  & 534 & 221  & 307 & 221 & 86 & - & - & - \\
    \hline
    \textbf{Car} & 10518  & 7520  & 2998  & 2332 & 1676 & 656 & - & - & -  \\
    \hline
    \textbf{Motorbike} & 83  & 61  & 22   & - & - & - & - & - & - \\
    \hline
    \textbf{Person} & 990  & 792  & 198  & - & - & - & 6525 & 5220 & 1305 \\
    \hline
    \textbf{Other} & 123 & 89  & 34  & 204 & 157 & 47 & - & - & -  \\
    \hline
    \textbf{Boat} & 498  & 349  & 149  & 171 & 120 & 51 & - & - & -  \\
    \hline
    \textbf{Bus} & 268  & 194  & 74  & - & - & - & - & - & - \\
    \hline
    \textbf{Tractor} & -  & -  & -  & 190 &  137 &  53 & - & - & -  \\
    \hline
    \textbf{Plane} & -  & -  & -  & 48 & 36 &  12 & - & - & -  \\
    \hline
    \textbf{Number of GTs} & 15247 & 10976 & 4271  & 3750 & 2697 & 1053 & 6525 & 5220 & 1305  \\
    \hline
    \textbf{Number of Images} & 1275 & 898 & 377  & 1246 & 892 & 354 & 1980 & 1601  & 379  \\
    \hline
    \end{tabular}%
    \vspace{0.2cm}
    \caption{\label{tab:statistics_dataset}Number of GT objects in the global, training and test sets for each class for AIRES (left), VEDAI (center) and SARD (right) datasets.} 
\end{table*}%

\begin{figure*}[!tb]
\centering
	\vspace{0cm}
	\includegraphics[width= \textwidth, keepaspectratio]{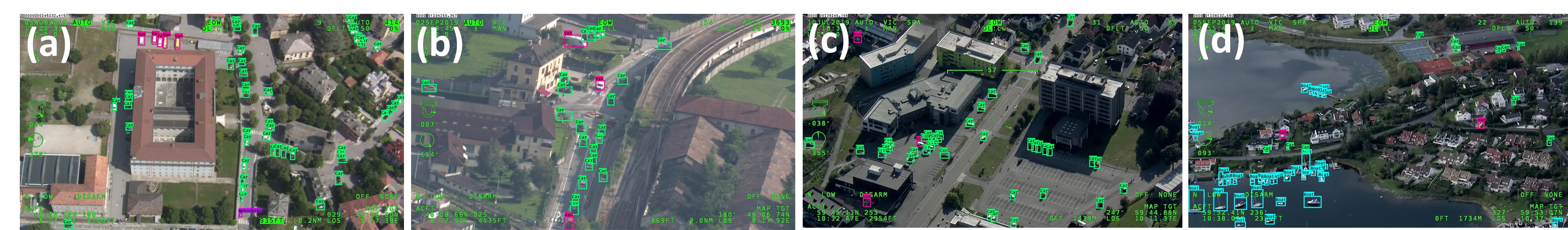}
	\vspace{-0.4cm}
	\caption{Some images of the AIRES dataset: (a) and (b) have been collected in Italy, whereas (c) and (d) in Norway. The vehicles are delimited by GT bounding boxes.} 
	\label{fig:examples_dataset}
\end{figure*}

As shown in Fig.~\ref{fig:examples_dataset}a, the images are characterized by variability of target dimension due to image acquisition at different altitudes and multiple viewpoints, as well as by the presence of different backgrounds (urban or rural areas), lighting conditions and target occlusion, blurring and haze. 

\begin{figure}[!tb]
\centering
	\vspace{0cm}
	\includegraphics[width= 9.0cm, keepaspectratio,trim={0.7cm 0 0 0},clip=true]{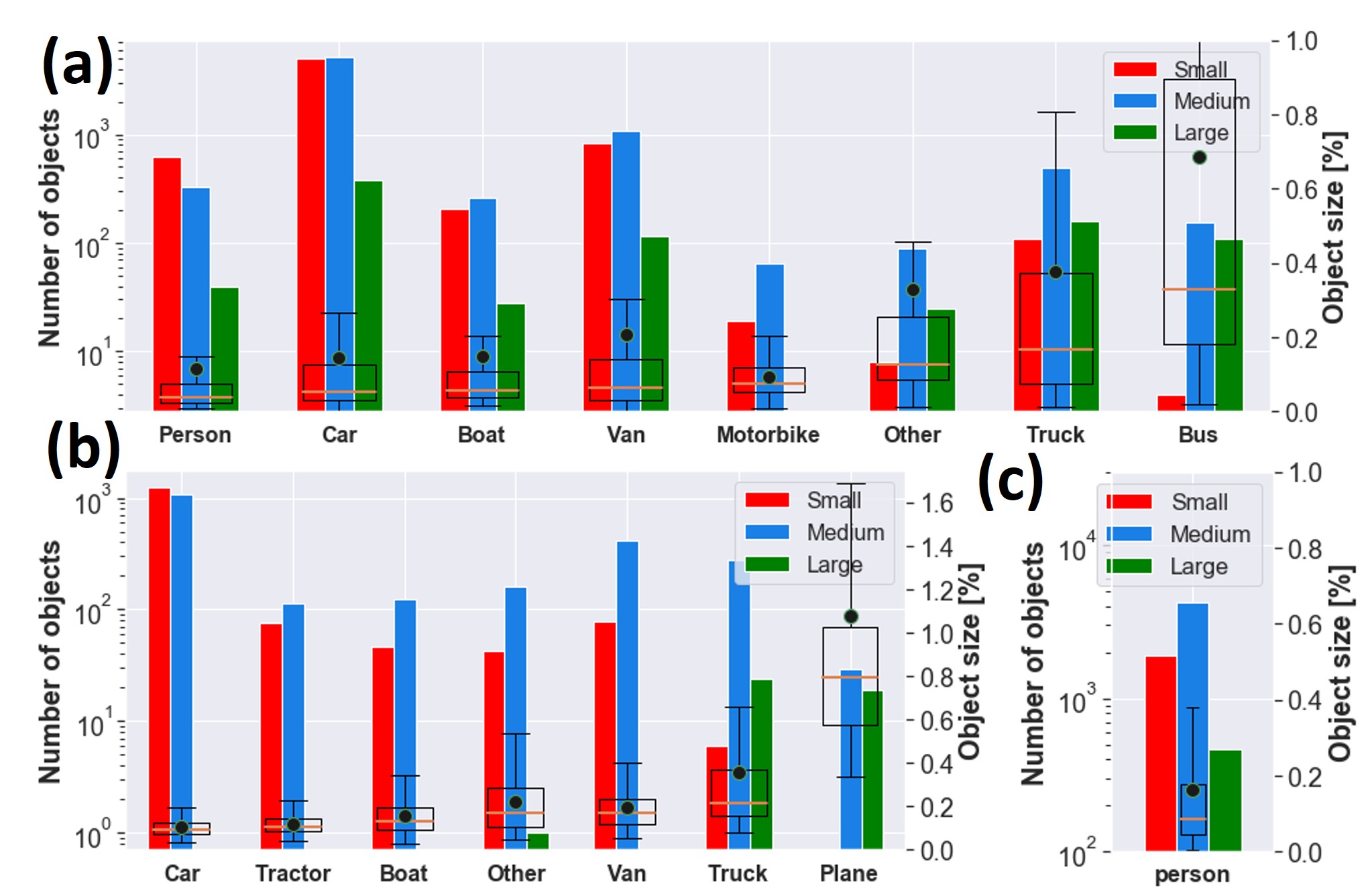}
	\vspace{-0.4cm}
	\caption{Distribution of GT objects for (a) AIRES, (b) VEDAI  and (c) SARD datasets. Left plot: histogram of small, medium and large objects grouped by class according to COCO convention~\cite{Lin_2014}. Right plot: box plot with 25th, 50th and 75th percentiles of target area for each class expressed as a percentage of the image size. The mean area is also shown as a black circle.} 
	\label{fig:area_coco}
\end{figure}

According to COCO's convention, objects can be categorized in "small" (area $<$ 32\textsuperscript{2}), "medium"  (32\textsuperscript{2} $\leq$ area $<$ 96\textsuperscript{2}) and "large" (area $\geq$ 96\textsuperscript{2}). 
Fig.~\ref{fig:area_coco}a shows the frequency for each category, as well as the object size. 
Medium targets are usually the most representative size for each class. Small objects are widely present as well for "Car", with almost 5k instances, "Van", "Person", "Boat" and "Truck". Large vehicles are instead statistically significant especially for categories "Bus", "Truck" and "Other".     
The largest targets are "Bus" and "Truck", with a median area of almost 0.33\% and 0.17\%, i.e. 83$\times$83 and 59$\times$59 pixel, respectively, and also a large size variance. The smallest instances are instead "Person", with a median size of almost 0.04\% or 29$\times$29 pixel, as well as "Car" and "Boat", both covering 0.05\% of the image size, i.e. roughly 32$\times$32 pixel.

\begin{figure*}[!tb]
\centering
	\vspace{0cm}
	\includegraphics[width= \textwidth, keepaspectratio]{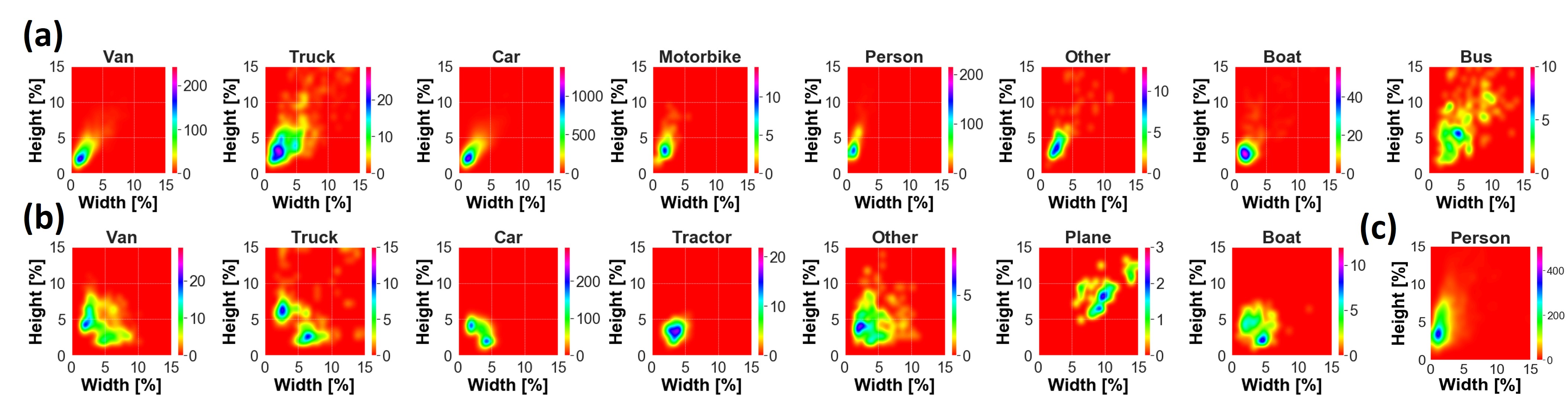}
	\vspace{-0.4cm}
	\caption{2D density plot of ground truth objects in the plane (GT width, GT height) for (a) AIRES,  (b) VEDAI and (c) SARD datasets. GT size is normalized with respect to the image size. The shape of source images are 1920$\times$1080 for (a) and (c) and 1024$\times$1024 for (b), respectively.} 
	\label{fig:heatmap_width_height}
\end{figure*}

In Fig.~\ref{fig:heatmap_width_height} we explore the 2D density distribution of ground truth objects in the plane defined by the target width and height, both being expressed as a percentage of the image size. 
Samples are mainly crowded in the region corresponding to width and height smaller than a few percent, except mostly for classes "Bus", "Truck" and "Other", where a remarkable presence of objects up to 10\%-15\% occurs.

\subsection{DOTAv2 and VEDAI datasets}\label{Dataset_others}
In our experiments we also used the publicly available vehicle datasets DOTAv2~\cite{DOTAv2_url} and VEDAI~\cite{VEDAI_url}.
VEDAI is a dataset of 1246 aerial images 1024$\times$1024 cropped from satellite images of the Utah Automated Geographic Reference Center (AGRC)~\cite{Utah_url}. It includes 12 classes, namely "car", "truck", "pickup", "tractor", "camping\_car", "boat", "motorcycle", "bus", "van", "other", "board", "plane". 
To make consistent VEDAI's classification with AIRES dataset, we merged "car" and "pickup" as one class "car", and "camping\_car" and "van" as "van". We also exploited the categories: "truck", "boat", "other", "tractor" and "plane" (Tab.~\ref{tab:statistics_dataset}). Globally, 3750 ground truths were used, with "car" representing roughly the 62.2\% of the overall statistics and "plane" just the 1.3\%.   
Fig.~\ref{fig:area_coco}b shows the distribution of objects for each class on the source images 1024$\times$1024. Medium sized objects are predominant for all categories, but "car", for which  small and medium groups are almost balanced and include more than one thousand of instances each. Large objects statistics is instead under-represented and mainly available only for "plane" and "truck". As can be seen in Figs.~\ref{fig:area_coco}-\ref{fig:heatmap_width_height}b, the median size of the targets is concentrated in the range from 0.09\% for "car" (31$\times$31 pixel), up to roughly 0.22\% and 0.80\% for "truck" and "plane", i.e. 48$\times$48 and 92$\times$92 pixel, respectively.

DOTAv2 is a large-scale dataset for object detection composed by satellite images collected by Google Earth, GF-2 and JL-1, as well as by aerial images provided by CycloMedia. It contains 2421 annotated training/validation images with variable size from a few Megapixel up to more than 100 Megapixel. Eighteen categories are present, including also very large targets such as "airport", "bridge"  and "soccer-ball-field". 
In this case we selected "small-vehicle", "large-vehicle", "ship" and "plane", where the first three classes were renamed as "car", "truck" and "boat", respectively. Globally, 314903 targets have been adopted for model learning.

\subsection{The Search and Rescue dataset (SARD) for person detection}\label{Dataset_SARD}
To further verify the generalization performances of the CNNs in a different real-world scenario, we also set up an experiment on SARD dataset~\cite{Sambolek_2021}, which is a recent collection of 1980 FHD aerial images 1920$\times$1080 extracted from a 35-minutes long video acquired from UAV and built for search and rescue operation (SAR) of people in distress or danger and available at the link~\cite{SARD_url}. The frames show actors simulating standard positions (standing, sitting, walking, running, lying), as well as positions typical of exhausted or injured persons. Further, they include manifold backgrounds such as woodland edge, low and high grass, lake basin, quarries and macadam roads.
The persons are manually tagged with tight bounding boxes: globally, the dataset contains 1980 images and 6525 instances of class "person" (Tab.~\ref{tab:statistics_dataset}). The median size area is less than 0.1\% of the full-sized image (Fig.~\ref{fig:area_coco}c), which corresponds to less than 45$\times$45 pixel. Medium targets are predominant, with almost one third of small objects and a few hundred large targets. Persons are more elongated vertically than horizontally, denoting a running or walking or standing posture, with a height up to roughly 10\% (Figs.~\ref{fig:heatmap_width_height}c).

\section{Methods}\label{Methods}
\subsection{The proposed networks YOLO-L and YOLO-S}\label{Methods_CNN}

\begin{figure*}[!tb]
\centering
	\vspace{0cm}
	\includegraphics[width= \textwidth, keepaspectratio]{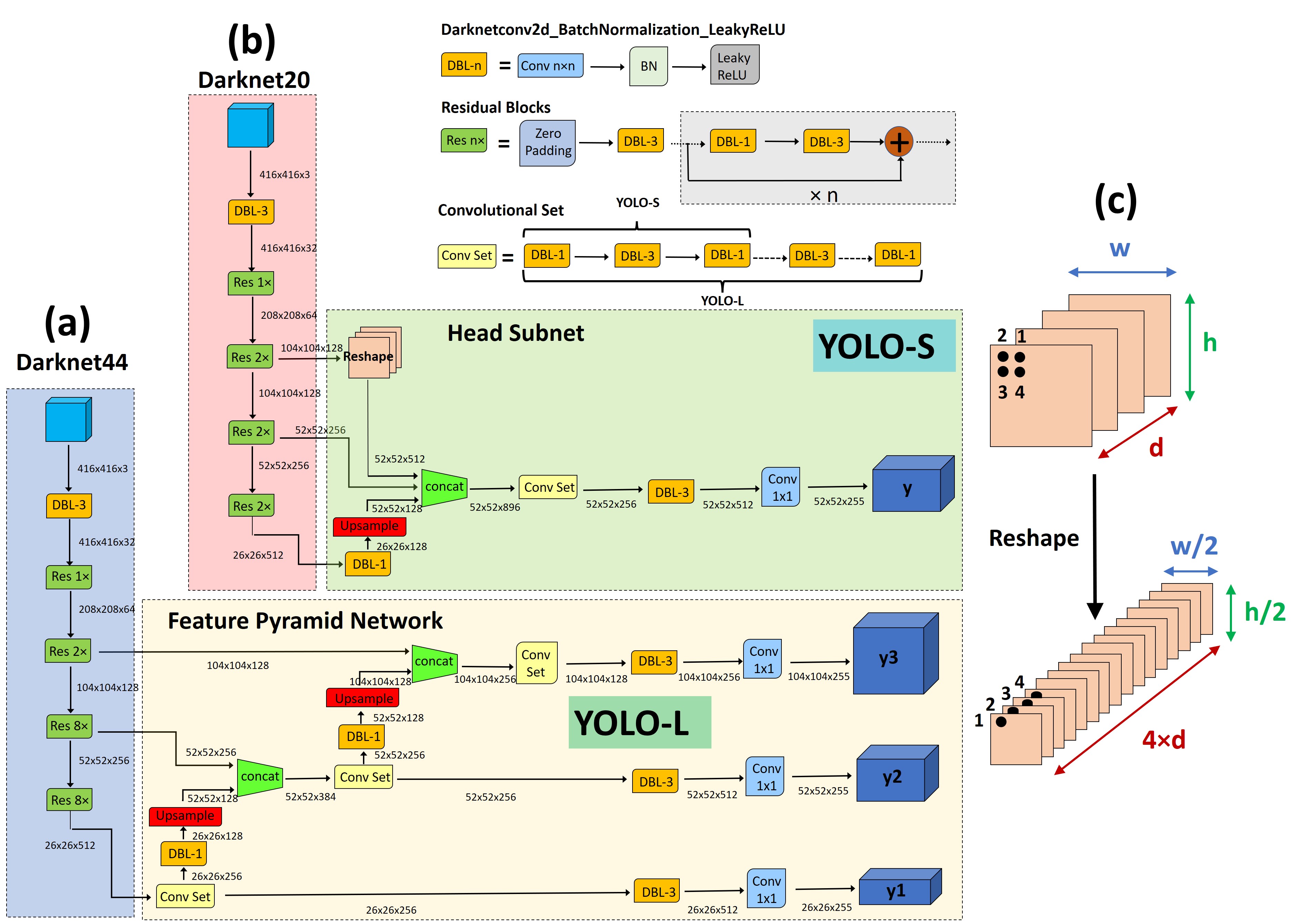}
	\vspace{0cm}
	\caption{Overview of the proposed networks: (a) YOLO-L, (b) YOLO-S, (c) Reshape - Passthrough layer. In the figure $C=80$ classes have been assumed (dataset COCO~\cite{Lin_2014}).} 
	\label{fig:networks}
\end{figure*}

In this work we propose two novel YOLO-like architectures specifically designed to meet the request of small target detection: YOLO-L and YOLO-S, depicted in Fig.~\ref{fig:networks}. 
YOLO-L, namely YOLO-\emph{large}, is proposed mainly for benchmarking purpose due to the limited inference speed which makes it suitable only for offline data processing on high-power hardware. YOLO-S, or YOLO-\emph{small}, is instead our proposal for an efficient, lightweight and accurate network to be deployed on edge devices.

Full details about the proposed CNNs are available in Tab.~\ref{tab:architectures}, where we report also the receptive field and cumulative stride for each layer. 
We assume an input image resized to the default size 416$\times$416. 
Further information about networks size, billion floating point operations (BFLOPs) required for each forward pass and properties of output layers are instead provided in Tab.~\ref{tab:cnn_properties}, where the proposed networks are compared to other state of the art detectors.

Receptive field $RF_k$ of layer $k$ has been computed according to the formula:
\begin{equation}
RF_k= RF_{k-1} + \left(f_k - 1 \right) \times CS_{k-1} \, ,
\end{equation}
where 
\begin{equation}
CS_{k-1}= \prod_{i=1}^{k-1} s_i 
\end{equation}
is the cumulative stride of layer $k-1$, $f_k$ is the filter size of layer $k$ and $s_i$ is the stride of layer $i$, respectively. For upsampling layers we can assume a kernel size equal to the number of input features involved in the computation of an output feature. As a consequence, if upsampling layer $k$ doubles the feature map size, $f_k= 1$ and $RF_k= RF_{k-1}$.


The larger network YOLO-L (Fig.~\ref{fig:networks}a) employs Darknet44 as backbone and a head subnet based on a Feature Pyramid Network~\cite{Lin_2016} to make detection at three different scales: 26$\times$26, 52$\times$52 and 104$\times$104. The rationale behind this choice is the fact that low-resolution feature maps contain grid cells that cover larger regions of the input space or, equivalently, have a large receptive field, and are therefore more suitable for detecting larger objects. On the contrary, grid cells from higher resolution feature maps look at a smaller portion of the input image and are thus better for detecting smaller objects. As a consequence, in order to make the network more focused on small targets, we decrease the receptive field of YOLO-L with respect to YOLOv3 by replacing the coarse-grained YOLOv3’s output scale 13$\times$13 with a finer-grained 4$\times$ down-sampled feature map of size 104$\times$104. 
In this way, YOLO-L shares two receptive fields with YOLOv3, whereas the largest receptive field 917$\times$917 of YOLOv3’s output scale 13$\times$13 is replaced by the much smaller 53$\times$53 (Tab.~\ref{tab:cnn_properties}).

In addition, a smaller backbone is also adopted, i.e. Darknet44, removing the last 4 residual blocks of Darknet53 corresponding to 32$\times$ down-sampled layers, which were used in YOLOv3 to downscale the image down to 13$\times$13. This has also the advantage to increase the processing speed. Globally, YOLO-L includes almost 23.848 millions of parameters, i.e. roughly 38.7\% of YOLOv3’s weights and requires 90.53 BFLOPs for each input image (Tab.~\ref{tab:cnn_properties}).

YOLO-S (Fig.~\ref{fig:networks}b) is instead a tiny and quick network that exploits the concepts of feature fusion and reshape - passthrough layer (Fig.~\ref{fig:networks}c) to combine precise location information of earlier fine-grained feature maps with more meaningful semantic information from deeper feature maps having lower resolution. Basically, it is based on a Darknet20 backbone, replacing the max pooling layers of Tiny-YOLOv3 in the feature-extraction stage with strided convolutional layers and residual units to reduce information loss during downsampling and increase efficiently the receptive field. The lightweight backbone, composed by seven residual blocks, allows also to avoid useless convolution operations for small-scale detected objects, which otherwise in a deeper architecture could lead to final features with a few pixels left after multiple down-sampling. 
In addition, YOLO-S employs a head subnet with one single output scale 52$\times$52 and a smaller convolutional set of just 4 alternate convolutional layers 1$\times$1 and 3$\times$3, instead of 6 as in YOLO-L and YOLOv3, to speed up inference. This leads to a receptive field of the output as large as 101$\times$101, sufficient to get meaningful contextual information around targets once source images are rescaled to the size expected by the network. 

Finally skip connection is implemented to extract more robust localization features by laterally connecting the eighth, thirteenth and nineteenth layers of the backbone, corresponding to 4$\times$, 8$\times$ and 16$\times$ down-sampled feature maps, respectively. 
Since such feature maps exhibit different resolutions, upsampling (reshaping) is applied to the nineteenth (eighth) layer to match every size to the shape 52$\times$52 before concatenation. Specifically, the reshape - passthrough layer keeps constant the number of input features by transforming from $d$ feature maps of size w$\times$h to $4 \cdot d$ feature maps with dimension w/2$\times$h/2 and thus making possible merging with deeper layers.
Overall, YOLO-S has a model volume shrinked by 87\% with respect to YOLOv3 and contains almost 7.853 millions of trainable parameters, resulting even lighter than Tiny-YOLOv3. Also, it requires 34.59 BFLOPs, which is close to SlimYOLOv3-SPP3-50~\cite{Zhang_2019} and almost one half of YOLOv3 (Tab.~\ref{tab:cnn_properties}). 

\begin{table*}[!htb]
\tiny
  \centering
    \begin{tabular}{p{0.1cm}p{1.2cm}p{0.3cm}p{0.3cm}p{1.2cm}p{1.2cm}p{0.2cm}p{0.2cm}|p{1.2cm}p{0.3cm}p{0.3cm}p{1.2cm}p{1.2cm}p{0.2cm}p{0.2cm}}
    \hline
    & & & &\textbf{YOLO-L}& & &                           & & & & \textbf{YOLO-S} & & & \\
    \hline
    \textbf{\#} & \textbf{Type} & \textbf{F} & \textbf{S/S} & \textbf{Input} & \textbf{Output} & \textbf{CS} & \textbf{RF} & \textbf{Type} & \textbf{F} & \textbf{S/S} & \textbf{Input} & \textbf{Output} & \textbf{CS} & \textbf{RF} \\
\hline
    0     & Conv  & 32    & 3/1   & 416$\times$416$\times$3 & 416$\times$416$\times$32 & 1     & 3     & Conv & 32    & 3/1 & 416$\times$416$\times$3 & 416$\times$416$\times$32 & 1     & 3 \\ \hline
    1     & Conv  & 64    & 3/2   & 416$\times$416$\times$32 & 208$\times$208$\times$64 & 2     & 5     & Conv & 64    & 3/2 & 416$\times$416$\times$32 & 208$\times$208$\times$64 & 2     & 5 \\ \hline
    2     & Conv (R1) & 32    & 1/1   & 208$\times$208$\times$64 & 208$\times$208$\times$32 & 2     & 5     & Conv (R1) & 32    & 1/1 & 208$\times$208$\times$64 & 208$\times$208$\times$32 & 2     & 5 \\ \hline
    3     & Conv (R1) & 64    & 3/1   & 208$\times$208$\times$32 & 208$\times$208$\times$64 & 2     & 9     & Conv (R1) & 64    & 3/1 & 208$\times$208$\times$32 & 208$\times$208$\times$64 & 2     & 9 \\ \hline
    4     & Conv  & 128   & 3/2   & 208$\times$208$\times$64 & 104$\times$104$\times$128 & 4     & 13    & Conv & 128   & 3/2 & 208$\times$208$\times$64 & 104$\times$104$\times$128 & 4     & 13 \\ \hline
    5     & Conv (R1) & 64    & 1/1   & 104$\times$104$\times$128 & 104$\times$104$\times$64 & 4     & 13    & Conv (R1) & 64    & 1/1 & 104$\times$104$\times$128 & 104$\times$104$\times$64 & 4     & 13 \\ \hline
    6     & Conv (R1) & 128   & 3/1   & 104$\times$104$\times$64 & 104$\times$104$\times$128 & 4     & 21    & Conv (R1) & 128   & 3/1 & 104$\times$104$\times$64 & 104$\times$104$\times$128 & 4     & 21 \\ \hline
    7     & Conv (R2) & 64    & 1/1   & 104$\times$104$\times$128 & 104$\times$104$\times$64 & 4     & 21    & Conv (R2) & 64    & 1/1 & 104$\times$104$\times$128 & 104$\times$104$\times$64 & 4     & 21 \\ \hline
    8     & Conv (R2) & 128   & 3/1   & 104$\times$104$\times$64 & 104$\times$104$\times$128 & 4     & 29    & Conv (R2) & 128   & 3/1 & 104$\times$104$\times$64 & 104$\times$104$\times$128 & 4     & 29 \\ \hline
    9     & Conv  & 256   & 3/2   & 104$\times$104$\times$128 & 52$\times$52$\times$256 & 8     & 37    & Conv & 256   & 3/2 & 104$\times$104$\times$128 & 52$\times$52$\times$256 & 8     & 37 \\ \hline
    10    & Conv (R1) & 128   & 1/1   & 52$\times$52$\times$256 & 52$\times$52$\times$128 & 8     & 37    & Conv (R1) & 128   & 1/1 & 52$\times$52$\times$256 & 52$\times$52$\times$128 & 8     & 37 \\ \hline
    11    & Conv (R1) & 256   & 3/1   & 52$\times$52$\times$128 & 52$\times$52$\times$256 & 8     & 53    & Conv (R1) & 256   & 3/1 & 52$\times$52$\times$128 & 52$\times$52$\times$256 & 8     & 53 \\ \hline
    12    & Conv (R2) & 128   & 1/1   & 52$\times$52$\times$256 & 52$\times$52$\times$128 & 8     & 53    & Conv (R2)& 128   & 1/1 & 52$\times$52$\times$256 & 52$\times$52$\times$128 & 8     & 53 \\ \hline
    13    & Conv (R2) & 256   & 3/1   & 52$\times$52$\times$128 & 52$\times$52$\times$256 & 8     & 69    & Conv (R2) & 256   & 3/1 & 52$\times$52$\times$128 & 52$\times$52$\times$256 & 8     & 69 \\ \hline
    14    & Conv (R3) & 128   & 1/1   & 52$\times$52$\times$256 & 52$\times$52$\times$128 & 8     & 69    & Conv & 512   & 3/2 & 52$\times$52$\times$256 & 26$\times$26$\times$512 & 16    & 85 \\ \hline
    15    & Conv (R3) & 256   & 3/1   & 52$\times$52$\times$128 & 52$\times$52$\times$256 & 8     & 85    & Conv (R1) & 256   & 1/1 & 26$\times$26$\times$512 & 26$\times$26$\times$256 & 16    & 85 \\ \hline
    16    & Conv (R4) & 128   & 1/1   & 52$\times$52$\times$256 & 52$\times$52$\times$128 & 8     & 85    & Conv (R1) & 512   & 3/1 & 26$\times$26$\times$256 & 26$\times$26$\times$512 & 16    & 117 \\ \hline
    17    & Conv (R4) & 256   & 3/1   & 52$\times$52$\times$128 & 52$\times$52$\times$256 & 8     & 101   & Conv (R2) & 256   & 1/1 & 26$\times$26$\times$512 & 26$\times$26$\times$256 & 16    & 117 \\ \hline
    18    & Conv (R5) & 128   & 1/1   & 52$\times$52$\times$256 & 52$\times$52$\times$128 & 8     & 101   & Conv (R2) & 512   & 3/1 & 26$\times$26$\times$256 & 26$\times$26$\times$512 & 16    & 149 \\ \hline
    19    & Conv (R5) & 256   & 3/1   & 52$\times$52$\times$128 & 52$\times$52$\times$256 & 8     & 117   & Conv & 128   & 1/1 & 26$\times$26$\times$512 & 26$\times$26$\times$128 & 16    & 149 \\ \hline
    20    & Conv (R6) & 128   & 1/1   & 52$\times$52$\times$256 & 52$\times$52$\times$128 & 8     & 117   & Upsample &       & 2/1 & 26$\times$26$\times$128 & 52$\times$52$\times$128 &  16   &  149 \\ \hline
    21    & Conv (R6) & 256   & 3/1   & 52$\times$52$\times$128 & 52$\times$52$\times$256 & 8     & 133   & Route 8 &       &       &       &       &       &  \\ \hline
    22    & Conv (R7) & 128   & 1/1   & 52$\times$52$\times$256 & 52$\times$52$\times$128 & 8     & 133   & Reshape &       &       & 104$\times$104$\times$128 & 52$\times$52$\times$512 &  4   &  29 \\ \hline
    23    & Conv (R7) & 256   & 3/1   & 52$\times$52$\times$128 & 52$\times$52$\times$256 & 8     & 149   & Route 22,20,13  &       &       &       &       &       &  \\ \hline
    24    & Conv (R8) & 128   & 1/1   & 52$\times$52$\times$256 & 52$\times$52$\times$128 & 8     & 149   & Conv & 256   & 1/1 & 52$\times$52$\times$896 & 52$\times$52$\times$256 & 8    & 69 \\ \hline
    25    & Conv (R8) & 256   & 3/1   & 52$\times$52$\times$128 & 52$\times$52$\times$256 & 8     & 165   & Conv & 512   & 3/1 & 52$\times$52$\times$256 & 52$\times$52$\times$512 & 8     & 85 \\ \hline
    26    & Conv  & 512   & 3/2   & 52$\times$52$\times$256 & 26$\times$26$\times$512 & 16    & 181   & Conv & 256   & 1/1 & 52$\times$52$\times$512 & 52$\times$52$\times$256 & 8     & 85 \\ \hline
    27    & Conv (R1) & 256   & 1/1   & 26$\times$26$\times$512 & 26$\times$26$\times$256 & 16    & 181   & Conv & 512   & 3/1 & 52$\times$52$\times$256 & 52$\times$52$\times$512 & 8     & 101 \\ \hline
    28    & Conv (R1) & 512   & 3/1   & 26$\times$26$\times$256 & 26$\times$26$\times$512 & 16    & 213   & Conv & 255   & 1/1 & 52$\times$52$\times$512 & 52$\times$52$\times$255 & 8     & 101 \\ \hline
    29    & Conv (R2) & 256   & 1/1   & 26$\times$26$\times$512 & 26$\times$26$\times$256 & 16    & 213   &   \textbf{Yolo}    &       &       &       &       &       &  \\ \hline
    30    & Conv (R2) & 512   & 3/1   & 26$\times$26$\times$256 & 26$\times$26$\times$512 & 16    & 245   &       &       &       &       &       &       &  \\ \hline
    31    & Conv (R3) & 256   & 1/1   & 26$\times$26$\times$512 & 26$\times$26$\times$256 & 16    & 245   &       &       &       &       &       &       &  \\ \hline
    32    & Conv (R3) & 512   & 3/1   & 26$\times$26$\times$256 & 26$\times$26$\times$512 & 16    & 277   &       &       &       &       &       &       &  \\ \hline
    33    & Conv (R4) & 256   & 1/1   & 26$\times$26$\times$512 & 26$\times$26$\times$256 & 16    & 277   &       &       &       &       &       &       &  \\ \hline
    34    & Conv (R4) & 512   & 3/1   & 26$\times$26$\times$256 & 26$\times$26$\times$512 & 16    & 309   &       &       &       &       &       &       &  \\ \hline
    35    & Conv (R5) & 256   & 1/1   & 26$\times$26$\times$512 & 26$\times$26$\times$256 & 16    & 309   &       &       &       &       &       &       &  \\ \hline
    36    & Conv (R5) & 512   & 3/1   & 26$\times$26$\times$256 & 26$\times$26$\times$512 & 16    & 341   &       &       &       &       &       &       &  \\ \hline
    37    & Conv (R6) & 256   & 1/1   & 26$\times$26$\times$512 & 26$\times$26$\times$256 & 16    & 341   &       &       &       &       &       &       &  \\ \hline
    38    & Conv (R6) & 512   & 3/1   & 26$\times$26$\times$256 & 26$\times$26$\times$512 & 16    & 373   &       &       &       &       &       &       &  \\ \hline
    39    & Conv (R7) & 256   & 1/1   & 26$\times$26$\times$512 & 26$\times$26$\times$256 & 16    & 373   &       &       &       &       &       &       &  \\ \hline
    40    & Conv (R7) & 512   & 3/1   & 26$\times$26$\times$256 & 26$\times$26$\times$512 & 16    & 405   &       &       &       &       &       &       &  \\ \hline
    41    & Conv (R8) & 256   & 1/1   & 26$\times$26$\times$512 & 26$\times$26$\times$256 & 16    & 405   &       &       &       &       &       &       &  \\ \hline
    42    & Conv (R8) & 512   & 3/1   & 26$\times$26$\times$256 & 26$\times$26$\times$512 & 16    & 437   &       &       &       &       &       &       &  \\ \hline
    43    & Conv  & 256   & 1/1   & 26$\times$26$\times$512 & 26$\times$26$\times$256 & 16    & 437   &       &       &       &       &       &       &  \\ \hline
    44    & Conv  & 512   & 3/1   & 26$\times$26$\times$256 & 26$\times$26$\times$512 & 16    & 469   &       &       &       &       &       &       &  \\ \hline
    45    & Conv  & 256   & 1/1   & 26$\times$26$\times$512 & 26$\times$26$\times$256 & 16    & 469   &       &       &       &       &       &       &  \\ \hline
    46    & Conv  & 512   & 3/1   & 26$\times$26$\times$256 & 26$\times$26$\times$512 & 16    & 501   &       &       &       &       &       &       &  \\ \hline
    47    & Conv  & 256   & 1/1   & 26$\times$26$\times$512 & 26$\times$26$\times$256 & 16    & 501   &       &       &       &       &       &       &  \\ \hline
    48    & Conv  & 512   & 3/1   & 26$\times$26$\times$256 & 26$\times$26$\times$512 & 16    & 533   &       &       &       &       &       &       &  \\ \hline
    49    & Conv  & 255   & 1/1   & 26$\times$26$\times$512 & 26$\times$26$\times$255 & 16    & 533   &       &       &       &       &       &       &  \\ \hline
    50    & \textbf{Yolo} &       & &  &  &       &       &       &       &       &       &       &       &  \\ \hline

    51    & Route 47 &       & &  &  &       &       &       &       &       &       &       &       &  \\ \hline
    52    & Conv  & 128   & 1/1   & 26$\times$26$\times$256 & 26$\times$26$\times$128 & 16    & 501   &       &       &       &       &       &       &  \\ \hline
    53    & Upsample &       & 2/1   & 26$\times$26$\times$128 & 52$\times$52$\times$128 &   16    &   501    &       &       &       &       &       &       &  \\ \hline
    54    & Route 53,25 &       &  & &  &       &       &       &       &       &       &       &       &  \\ \hline
    55    & Conv  & 256   & 1/1   & 52$\times$52$\times$384 & 52$\times$52$\times$256 & 8     & 165   &       &       &       &       &       &       &  \\ \hline
    56    & Conv  & 512   & 3/1   & 52$\times$52$\times$256 & 52$\times$52$\times$512 & 8     & 181   &       &       &       &       &       &       &  \\ \hline
    57    & Conv  & 256   & 1/1   & 52$\times$52$\times$512 & 52$\times$52$\times$256 & 8     & 181   &       &       &       &       &       &       &  \\ \hline
    58    & Conv  & 512   & 3/1   & 52$\times$52$\times$256 & 52$\times$52$\times$512 & 8     & 197   &       &       &       &       &       &       &  \\ \hline
    59    & Conv  & 256   & 1/1   & 52$\times$52$\times$512 & 52$\times$52$\times$256 & 8     & 197   &       &       &       &       &       &       &  \\ \hline
    60    & Conv  & 512   & 3/1   & 52$\times$52$\times$256 & 52$\times$52$\times$512 & 8     & 213   &       &       &       &       &       &       &  \\ \hline
    61    & Conv  & 255   & 1/1   & 52$\times$52$\times$512 & 52$\times$52$\times$255 & 8     & 213   &       &       &       &       &       &       &  \\ \hline
    62    & \textbf{Yolo} &       &  &  &  &       &       &       &       &       &       &       &       &  \\ \hline

    63    & Route 59 &       &  &  &  &       &       &       &       &       &       &       &       &  \\ \hline
    64    & Conv  & 128   & 1/1   & 52$\times$52$\times$256 & 52$\times$52$\times$128 & 8     & 197   &       &       &       &       &       &       &  \\ \hline
    65    & Upsample &       & 2/1   & 52$\times$52$\times$128 & 104$\times$104$\times$128 &   8    &  197     &       &       &       &       &       &       &  \\ \hline
    66    & Route 65,8 &       & &  &  &       &       &       &       &       &       &       &       &  \\ \hline
    67    & Conv  & 128   & 1/1   & 104$\times$104$\times$256 & 104$\times$104$\times$128 & 4     & 29    &       &       &       &       &       &       &  \\ \hline
    68    & Conv  & 256   & 3/1   & 104$\times$104$\times$128 & 104$\times$104$\times$256 & 4     & 37    &       &       &       &       &       &       &  \\ \hline
    69    & Conv  & 128   & 1/1   & 104$\times$104$\times$256 & 104$\times$104$\times$128 & 4     & 37    &       &       &       &       &       &       &  \\ \hline
    70    & Conv  & 256   & 3/1   & 104$\times$104$\times$128 & 104$\times$104$\times$256 & 4     & 45    &       &       &       &       &       &       &  \\ \hline
    71    & Conv  & 128   & 1/1   & 104$\times$104$\times$256 & 104$\times$104$\times$128 & 4     & 45    &       &       &       &       &       &       &  \\ \hline
    72    & Conv  & 256   & 3/1   & 104$\times$104$\times$128 & 104$\times$104$\times$256 & 4     & 53    &       &       &       &       &       &       &  \\ \hline
    73    & Conv  & 255   & 1/1   & 104$\times$104$\times$256 & 104$\times$104$\times$255 & 4     & 53    &       &       &       &       &       &       &  \\ \hline
    74    & \textbf{Yolo} &       & &  &  &       &       &       &       &       &       &       &       &  \\ \hline

    \hline
    \end{tabular}%
\vspace{0.2cm}
\caption{The architecture of the proposed networks YOLO-L (left) and YOLO-S (right). F:  number of convolutional filters; S/S: kernel size/stride; CS: cumulative stride; RF: receptive field; R: residual block. In the table $C=80$ classes have been assumed (dataset COCO~\cite{Lin_2014}).}
\label{tab:architectures}
  \label{tab:addlabel}%
\end{table*}%

\begin{table}[!htb]
\small
  \centering
    \begin{tabular}{ccccccc}
    \hline
    \textbf{Network} & \textbf{Parameters} & \textbf{Volume} & \textbf{BFLOPs} & \textbf{Output} & \textbf{RF} & \textbf{CS} \\
      &  &  & &\textbf{Scale} &  \\
    \hline
     & & & & 13$\times$13 & 917$\times$917 & 32\\
    \textbf{YOLOv3} & 61.603M & 242.9MB & 65.86 & 26$\times$26 & 533$\times$533 & 16 \\
     & & & & 52$\times$52 &  213$\times$213 & 8\\
	\hline    
    \textbf{Tiny-} & 8.688M & 34.7MB &  \textbf{5.58} & 13$\times$13 & 318$\times$318 & 32 \\
    \textbf{YOLOv3} & & & & 26$\times$26 & 110$\times$110 & 16\\
    \hline
    \textbf{M. Ju et} & \textbf{691.6K} & \textbf{3.4MB} & \textcolor{green}{\textbf{6.73}} & 52$\times$52 & 133$\times$133 & 8 \\
    \textbf{al~\cite{Ju_2019}} & & & &  &  \\
    \hline
    & & & & 26$\times$26 & 533$\times$533 & 16\\
    \textbf{YOLO-L} & 23.848M & 94.9MB & 90.53 & 52$\times$52 & 213$\times$213 & 8 \\
     & & & & 104$\times$104 & 53$\times$53 & 4\\
     \hline
     \textbf{YOLO-S} & \textcolor{green}{\textbf{7.853M}} &  \textcolor{green}{\textbf{31.9MB}} & 34.59 & 52$\times$52 & 101$\times$101 & 8 \\
	\hline
    \end{tabular}%
    \vspace{0.2cm}
    \caption{Number of parameters, volume, billion floating point operations (BFLOPs), output scales, receptive field (RF) and cumulative stride (CS) for each network. The lightest (2nd lightest) model is highlighted in black (green) bold. The two CNNs having smaller BFLOPs are also highlighted.}
  \label{tab:cnn_properties}%
\end{table}%

\subsection{State of the Art detectors}\label{Methods_CNN_other}
In this work we re-implemented the following State of the Art single-stage detectors for comparison with our networks: YOLOv3~\cite{Redmon_2018}, Tiny-YOLOv3~\cite{YOLO} and CNN proposed by Ref.~\cite{Ju_2019}. YOLOv3 exploits Darknet53 as feature extractor and adopts a head subnetwork based on a Feature Pyramid Network with three output scales from 13$\times$13 to 52$\times$52 to detect both large and small targets. 
It also implements features fusion via a top-down pathway and lateral connection between feature maps of different resolution to get accurate location information. 
Object detection is posed as a regression problem by resizing the input image to the default size 416$\times$416 and dividing it in a grid for each output scale. 
Each grid cell outputs an array with length $B\times(5+C)$, where $B$ is the
number of rectangular bounding boxes a cell can predict, 5 stands for the number of bounding
box attributes and the object confidence, and $C$ is the number of classes. Non Maximal
Suppression (NMS) is finally applied to preserve only the bounding boxes having highest confidence.
Overall, YOLOv3 includes 61.603 millions of parameters (Tab.~\ref{tab:cnn_properties}). 

With the aim of improving the detection speed and making actionable deploying on edge devices, Tiny-YOLOv3 is instead a light version with 5.58 BFLOPs and almost 8.688 millions of parameters obtained by replacing the feature extractor Darknet53 with a backbone composed by 7 convolutional and 6 max pooling layers (Tab.~\ref{tab:cnn_properties}). It makes detection at two different levels, i.e. 13$\times$13 and 26$\times$26, thus resulting less efficient in identifying small targets. 

Finally we introduce also the network presented recently by Ref.~\cite{Ju_2019}. Since it is fully convolutional, it can be fed with inputs of any size: assuming an image 416$\times$416, the output scale is equal to 52$\times$52, which corresponds to a receptive field as large as 133$\times$133. Additionally, the use of 1$\times$1 convolution to reduce channel dimension after each concatenation allows to obtain a very light CNN of almost 700 thousands of parameters, less than one tenth of Tiny-YOLOv3 (Tab.~\ref{tab:cnn_properties}). As already stated before, it exhibits an AP on VEDAI and DOTA datasets quite close to YOLOv3, but with a latency time very competitive with Tiny-YOLOv3 and almost one tenth BFLOPs of YOLOv3.

\begin{figure*}[!t]
\small
\centering
	\vspace{0cm}
	\includegraphics[width= 0.85\textwidth, trim={0cm 0 0 0}, clip ,keepaspectratio]{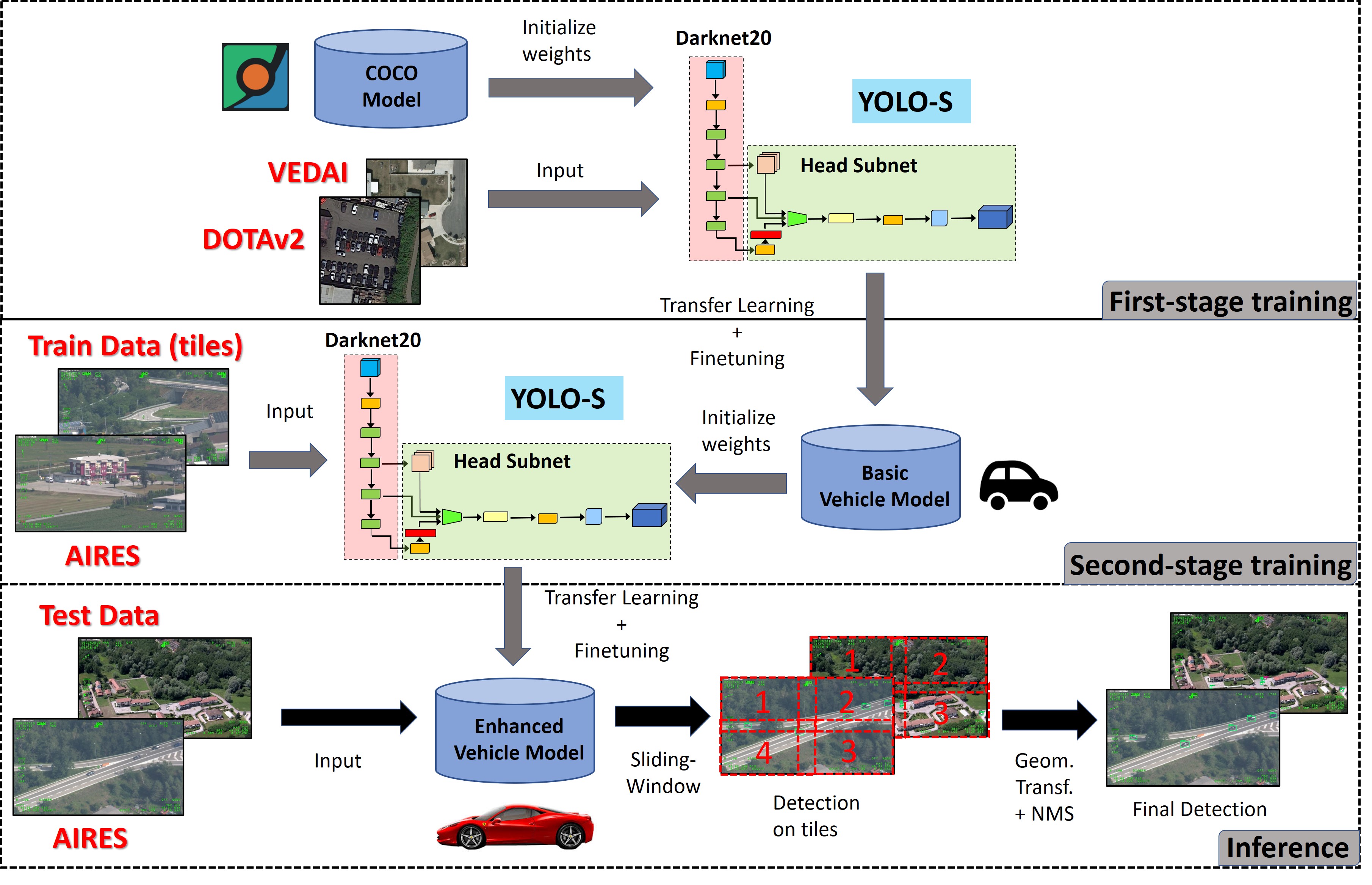}
	\vspace{0cm}
	\caption{Workflow of the proposed vehicle detection approach for experiment 1 on AIRES dataset.} 
	\label{fig:training_workflow}
\end{figure*}

\subsection{The experiments}\label{Methods_experiments}
In order to benchmark the proposed networks with baseline detectors, we defined a set of experiments on AIRES, VEDAI and SARD datasets. Further details are provided in the following sections for each dataset.

\subsubsection{Experiments on AIRES dataset}\label{Methods_experiments_our}
\textbf{Experiment 1}: evaluate the aforementioned CNNs on AIRES dataset by using a double-stage training and a sliding window approach during inference, as described below.

First, the dataset has been split randomly in training (70\%) and test (30\%) by implementing stratified sampling for each class (Tab.~\ref{tab:statistics_dataset}).  
Then, in order to enrich statistics available during model learning, we enabled standard data augmentation techniques including horizontal flipping, resizing, cropping and random distortion of brightness, contrast, saturation and hue.
However, the over-fitting issue due to a lack of data cannot often be resolved effectively with only data augmentation methods, especially for minority classes. 

As a consequence, we adopted the so called "transfer learning" technique in order to take advantage of knowledge achievable on publicly available databases present in Literature.  
This is especially beneficial for detection accuracy: the more similar the base task, on which preliminary features are extracted, to the target task of interest, the higher the accuracy achievable~\cite{Yosinski_2014}. 
The proposed training pipeline is illustrated in Fig.~\ref{fig:training_workflow}. Specifically, we implemented a double-stage training adding a transitional task between the source task based on COCO dataset and the target task on AIRES dataset: (i) first, a training on DOTAv2 $+$ VEDAI datasets and then (ii) on AIRES dataset. Pre-trained weights on COCO dataset were used as a warm starting point to speed up training and improve accuracy.

To make consistent VEDAI's data with AIRES dataset, during training phase (i) we used VEDAI statistics reported in Tab.~\ref{tab:statistics_dataset}, but excluding "plane" and "tractor". We also exploited all the DOTAv2 instances "car", "truck" and "boat" available.
As the size of DOTAV2 images were not uniform, we cropped every DOTAv2 and VEDAI images by means of an in-house algorithm in not-overlapping tiles having the minimum side not smaller than 416 pixels and retaining the original aspect ratio. Then we executed first-stage training on the resulting 74232 patches of joint training dataset DOTAv2 $+$ VEDAI by adding a lateral black padding on the smaller patch side, if necessary, to make image square as requested by YOLO-like networks. 

Second-stage training on AIRES dataset was then performed by starting from pre-trained weights achieved on DOTAv2 $+$ VEDAI. 
Training images were divided in patches in order to couple efficiently with the inference step. 
In particular, since image resizing from full-sized image W$\times$H to network size reduces a median-sized "Bus" object 83 pixel wide (see Fig.~\ref{fig:area_coco}a) to just 18 pixel, we implemented a sliding window approach during inference assuming $N_x= N_y= 2$ windows along both dimensions with an overlap $o_x= o_y= 50$ pixel, corresponding roughly to the third quartile of the overall population size. Specifically, the overlapping has been allowed to cope with detection of objects at the window edge and its value has been determined as a trade-off between the target localization accuracy and latency time, which increases proportionally to the overlapping area due to the post-processing NMS algorithm used to eliminate redundant detection occurrences over the same object in the overlapping regions after sliding and detection steps~\cite{Alganci_2020}. 

Hence, for consistency, we divided each input training image in four tiles of shape $\Delta_x \times \Delta_y=$ 985$\times$565 pixel, where tile size was computed as follows:
\begin{gather}
\Delta_x= \lceil \frac{W + \left(N_x - 1 \right)\cdot o_x}{N_x} \rceil \, , \\
\Delta_y= \lceil \frac{H + \left(N_y - 1 \right)\cdot o_y}{N_y} \rceil \, ,
\end{gather} 
where the operator $\lceil \cdot \rceil$ denotes the minimum integer larger or equal to its argument. 
During both training stages, we assumed a typical two-stage pipeline by first training only the last convolutional layers (transfer learning) and then unfreezing the whole network (finetuning). The rationale behind this choice is that earlier layers of the network typically learn low-level features useful in a broad range of vision tasks, such as edge, corner, shape and color, and thus should change less than deeper layers which are more specialized in the target task.
\\
\textbf{Experiment 2}: implement a single-stage training on AIRES dataset to verify the performance drop with respect to double-stage training. 

We adopted the same philosophy of Experiment 1, but using only pre-trained weights on COCO dataset. 
 
\subsubsection{Experiment on VEDAI dataset}\label{Methods_experiments_VEDAI}
\textbf{Experiment 1}: test the aforementioned CNNs on VEDAI dataset by using a double-stage training and a sliding window approach during inference.

As shown in Tab.~\ref{tab:statistics_dataset}, VEDAI has been split preliminarily in training (70\%) and test (30\%), whereas DOTAv2 has been used entirely for training. 
Then training has been modelled as follows: (i) preliminary training on DOTAv2 images, then (ii) on VEDAI dataset. Pre-trained weights on COCO were exploited as usual.
In addition, all VEDAI classes shown in Tab.~\ref{tab:statistics_dataset} were employed in this case. 
As in Section~\ref{Methods_experiments_our}, inference was executed by sliding a window over the whole input image with $N_x= N_y= 2$ and $o_x= o_y= 50$ pixel. 
Hence VEDAI training images were split accordingly in four tiles of size $\Delta_x \times \Delta_y=$ 537$\times$537 pixel.
\\
\textbf{Experiment 2}: implement a single-stage training on VEDAI dataset and infer over the full-sized image 1024$\times$1024 directly.

This method, which is the same applied in Ref.~\cite{Ju_2019}, aims to verify the robustness of the CNNs in a "single-shot" inference approach, i.e. when a more computationally demanding sliding window strategy is not a feasible solution due to time constraints. 
In this case we conducted a training on the whole VEDAI images directly based on pre-trained weights on COCO. 

\subsubsection{Experiment on SARD dataset}\label{Methods_experiment_SARD}
\textbf{Experiment}: implement a single-stage training on SARD dataset and infer over the full-sized image 1920$\times$1080.

The same approach of experiment 2 on VEDAI has been followed. In addition, a training-test splitting 80-20\% has been employed (Tab.~\ref{tab:statistics_dataset}).

\subsubsection{Evaluation metrics}\label{Methods_experiments_metrics}
In this work we introduce both the PASCAL VOC~\cite{Everingham_2015} and COCO~\cite{Lin_2014} evaluation protocols.
Performances are evaluated for each vehicle class in terms of  the  Average  Precision 
(AP), defined as the area under the Precision-Recall curve, i.e.
\begin{equation}
AP= \int_0^1 PREC(REC) \, dREC \, ,
\end{equation}
for different Intersection over Union (IoU) between the ground-truth box and the predicted one. Overall performance is computed by means of mAP, defined as:
\begin{equation}
mAP= \frac{1}{n} \cdot \sum_{i=1}^n AP_i \, ,
\end{equation}
where the sum runs over the $n$ object categories. We also introduce the weighted Average Precision (wAP), where the $i$-th class is weighted according to its support $N_i$ ($i$= 1,.., n):
\begin{equation}
wAP= \sum_{i=1}^n \frac{N_i}{N} \cdot AP_i 
\end{equation}
and $N= \sum_{i=1}^n N_i$ is the overall number of instances.
Additionally, we compute COCO metrics AP\textsubscript{S}, AP\textsubscript{M} and AP\textsubscript{L} by applying AP@[0.5:0.05:0.95], i.e. by computing the average of the APs evaluated at 10 different IoU thresholds, and considering only ground truth objects of small, medium and large size, respectively~\cite{Padilla_2021}. 

Furthermore, as datasets usually exhibit class imbalance, we compute the micro-average (ma) recall, precision and F1-score by aggregating contributions of all classes: 
\begin{gather} 
REC_{ma}= \frac{TP_{ma}}{P_{ma}} \\
PREC_{ma}= \frac{TP_{ma}}{TP_{ma} + FP_{ma}} \\
F1_{ma}= \frac{2 \times REC_{ma} \times PREC_{ma}}{REC_{ma} + PREC_{ma}}
\end{gather}
where TP\textsubscript{ma}, FP\textsubscript{ma} and P\textsubscript{ma} denote the overall number of true positives, false positives and positive samples in the dataset. 
Finally we estimate the processing speed based on the FPS processed by the different algorithms.   

\begin{table*}[!b]
\small
  \centering
    \begin{tabular}{cccccc}
    \hline
    \textbf{Metric} & \textbf{YOLOv3} & \textbf{Tiny-YOLOv3} & \textbf{M. Ju et al~\cite{Ju_2019}} & \textbf{YOLO-L} & \textbf{YOLO-S} \\
    \hline
     \textbf{Van [\%]} & 33.1  & 9.1  & 18.4   & \textcolor{green}{\textbf{36.6}}  &  \textbf{44.0} \\
     \textbf{Truck [\%]} & 47.1 & 17.5   &  17.0   & \textcolor{green}{\textbf{47.3}} &  \textbf{50.5} \\
     \textbf{Car [\%]} & 47.8  & 19.1   &  46.5   &  \textcolor{green}{\textbf{57.3}} & \textbf{58.7} \\
     \textbf{Motorbike [\%]} & \textcolor{green}{\textbf{20.4}}  &  11.4  & 0.0  & \textbf{26.6} & 13.6 \\
     \textbf{Person [\%]} & 35.5 & 6.4  &  15.2   & \textbf{47.6} & \textcolor{green}{\textbf{42.4}} \\
     \textbf{Other [\%]} & \textcolor{green}{\textbf{39.7}} & 6.5   &  0.0   & 20.3 & \textbf{39.9} \\
     \textbf{Boat [\%]} & 39.5  & 13.5   &  44.8   & \textcolor{green}{\textbf{52.5}}  & \textbf{65.9} \\
     \textbf{Bus [\%]} & \textcolor{green}{\textbf{57.7}} & 28.2  &  24.3   & \textbf{58.9}  & \textbf{58.9} \\
     \hline     
    \textbf{mAP [\%]} & 40.1  &  13.9   & 20.8    & \textcolor{green}{\textbf{43.4}} & \textbf{46.7} \\
     \hline
     \textbf{wAP [\%]} & 44.9	& 16.9	& 38.7	& \textcolor{green}{\textbf{52.9}}	& \textbf{55.4}
 \\
     \hline
    \textbf{AP\textsubscript{S} [\%]} & 5.5  & 0.8   & 3.8   & \textbf{11.9} & \textcolor{green}{\textbf{10.4}} \\
	\hline   
	 \textbf{AP\textsubscript{M} [\%]} & 17.5  &  5.1  &  11.3 & \textcolor{green}{\textbf{21.6}} &  \textbf{25.4} \\
	\hline  
	 \textbf{AP\textsubscript{L} [\%]} & \textcolor{green}{\textbf{28.1}}  &  11.1  & 10.9 & \textbf{29.3} & 22.1 \\
	\hline    
    \textbf{REC\textsubscript{ma} [\%]} & 54.4  & 27.5 & 46.4  &  \textcolor{green}{\textbf{60.7}} & \textbf{62.0} \\
	\hline    
    \textbf{PREC\textsubscript{ma} [\%]} & 62.9  & 38.6  & 63.4  & \textbf{71.4} & \textcolor{green}{\textbf{69.5}} \\
	\hline   
    \textbf{F1\textsubscript{ma} [\%]} & 58.3  & 32.1  & 53.6  & \textbf{65.6} & \textcolor{green}{\textbf{65.5}} \\
	\hline    
    \textbf{FPS} & 5.9   & \textbf{10.5} & \textcolor{green}{\textbf{9.7}} & 5.8   & 8.1 \\
    \hline
    \end{tabular}%
    \vspace{0.2cm}
    \caption{Experiment 1 on AIRES dataset. Comparative results of YOLOv3~\cite{Redmon_2018}, Tiny-YOLOv3~\cite{YOLO}, CNN by Ref.~\cite{Ju_2019}, YOLO-L and YOLO-S in terms of AP, mAP, wAP, AP\textsubscript{S}, AP\textsubscript{M}, AP\textsubscript{L}, REC\textsubscript{ma}, PREC\textsubscript{ma}, F1\textsubscript{ma} and speed (FPS). The 1st (2nd) best score is highlighted in black (green) bold.}
  \label{tab:performance_our_dataset}%
\end{table*}%

\section{Results}\label{Results}
All networks have been implemented in the same environment based on the Keras library within the TensorFlow framework. Due to the large number of simulations, experiments were conducted using two processors: an Intel(R) Core(TM) i7-4770 CPU @ 3.40GHz  with 16 GB RAM and NVIDIA QUADRO RTX 5000 GPU used for both training and inference, and an Intel Xeon E5-2690 v3 with 56 GB RAM and NVIDIA Tesla K80 GPU exploited only for training. 

All algorithms adopt the same objective function based on YOLOv3, which has been minimized during training according to the stochastic Adam optimizer. Furthermore a step-wise learning rate has been employed, starting from $10^{-3}$ and decreasing by a factor 10 after every 10 epochs without any loss improvement. Batch size was fixed to 32 during transfer learning and 8 during finetuning. Early stopping method was also used to avoid overfitting on training data by randomly leaving out 10\% of training instances for validation purpose. 
The network by Ref.~\cite{Ju_2019}, as well as YOLO-L and YOLO-S were preliminarily trained from scratch on COCO dataset for more than one week each, whereas for YOLOV3 and Tiny-YOLOv3 we exploited the pre-trained weights on COCO already available online~\cite{YOLO}.
Then the models were fine-tuned on the considered datasets for up to 1000 epochs: each training session took from one to some days. 

We estimated anchor priors based on an enhanced k-means clustering by replacing the euclidean distance with the IoU metric to avoid an error biased towards large bounding boxes, following for example Ref.~\cite{He_2019}. We notice that in Ref.~\cite{Ju_2019} details were not provided about the method employed to compute anchors, neither their number, for a sake of reproducibility. In this work 3 anchors have been assumed for each output scale of YOLOv3, Tiny-YOLOv3 and YOLO-L and 6 anchors for CNN by Ref.~\cite{Ju_2019} and YOLO-S, respectively. 
All the presented results, but COCO metrics, correspond to an intermediate IoU equal to 0.5.

\begin{table*}[!b]
\small
  \centering
    \begin{tabular}{ccccccc}
    \hline
    \textbf{Metric} & \textbf{Training} & \textbf{YOLOv3} & \textbf{Tiny-YOLOv3} & \textbf{M. Ju et al~\cite{Ju_2019}} & \textbf{YOLO-L} & \textbf{YOLO-S} \\
    \hline
    \textbf{mAP [\%]} & single-stage & \textbf{41.9}  & 10.9  & 20.3 & \textbf{47.3} &  44.3 \\
      \textbf{mAP [\%]} & double-stage & 40.1  &  \textbf{13.9}   & \textbf{20.8}    & 43.4 & \textbf{46.7} \\
      \hline
     \textbf{wAP [\%]} & single-stage & 43.2 &	11.0  &	34.5 &	52.0 &	53.1 \\
     \textbf{wAP [\%]} & double-stage & \textbf{44.9}	& \textbf{16.9}	& \textbf{38.7}	& \textbf{52.9}	& \textbf{55.4}
 \\
      \hline
            \textbf{AP\textsubscript{S} [\%]} & single-stage & \textbf{6.8}  & 0.4  & 2.8   & 11.5 & 9.4 \\
      \textbf{AP\textsubscript{S} [\%]} & double-stage & 5.5  & \textbf{0.8}  & \textbf{3.8} & \textbf{11.9} & \textbf{10.4} \\
	\hline   
	  \textbf{AP\textsubscript{M} [\%]} & single-stage & 17.4  &  3.1 & \textbf{11.4} & \textbf{22.4} & 23.7 \\
	 \textbf{AP\textsubscript{M} [\%]} & double-stage & \textbf{17.5}  &  \textbf{5.1}  &  11.3 & 21.6 & \textbf{25.4} \\
	\hline  
\textbf{AP\textsubscript{L} [\%]} & single-stage & \textbf{31.7}  &  10.5   &  10.2 & \textbf{31.0} & \textbf{22.8} \\
	\textbf{AP\textsubscript{L} [\%]} & double-stage & 28.1  &  \textbf{11.1}  & \textbf{10.9} & 29.3 & 22.1 \\
	\hline  
	 \textbf{REC\textsubscript{ma} [\%]} &  single-stage & 52.9 & 20.1  & 42.0  & 59.7  & 60.2 \\
	  \textbf{REC\textsubscript{ma} [\%]} &  double-stage & \textbf{54.4}  & \textbf{27.5}  & \textbf{46.4}  &  \textbf{60.7} & \textbf{62.0} \\
	\hline    
    \textbf{PREC\textsubscript{ma} [\%]} &  single-stage & 62.6  & 30.9 & 60.9  & 69.6 & 67.3 \\
     \textbf{PREC\textsubscript{ma} [\%]} &  double-stage & \textbf{62.9}  & \textbf{38.6}  & \textbf{63.4}  & \textbf{71.4} & \textbf{69.5} \\
     \hline
	    \textbf{F1\textsubscript{ma} [\%]} & single-stage & 57.3  & 24.3  & 49.7 & 64.3 &  63.6 \\
    \textbf{F1\textsubscript{ma} [\%]} & double-stage & \textbf{58.3}  & \textbf{32.1}  & \textbf{53.6}  & \textbf{65.6} & \textbf{65.5} \\
	\hline    
    \end{tabular}%
    \vspace{0.2cm}
    \caption{Experiment 2 on AIRES dataset. Metrics mAP, wAP, AP\textsubscript{S}, AP\textsubscript{M}, AP\textsubscript{L}, REC\textsubscript{ma}, PREC\textsubscript{ma} and F1\textsubscript{ma} for single-stage (experiment 2) and double-stage training (experiment 1). The best score is highlighted in black bold.}
  \label{tab:single_double_stage_training}%
\end{table*}%

\subsection{Experiments on AIRES dataset}

Due to the large size of test images, inference has been performed by moving a sliding window 985$\times$565 pixel across the image with a step size $\Delta_i - o_i$ of 935 (515) pixel along $i=x$ ($i=y$) direction. Outcomes have been finally recombined into the original input image by means of geometrical transformations and pruned according to NMS algorithm applied to the 50 pixel-wide overlapping regions.

\subsubsection{Quantitative comparison with State of the Art detectors}\label{Results_quantitative}

Performances for experiment 1 are shown in Tab.~\ref{tab:performance_our_dataset}. It may be seen that the proposed CNNs outperform all the other detectors, with a mAP of 46.7\% for YOLO-S, 43.4\% for YOLO-L, 40.1\% for YOLOv3, 20.8\% for~\cite{Ju_2019} and less than 14\% for Tiny-YOLOv3. In general, the larger the statistics available, the higher the AP achievable, with remarkable (modest) values for "Car" ("Motorbike", "Other"). In addition, despite of the low "Boat" statistics, AP larger than 50\% are obtained by the proposed CNNs, which may be explained by the broad receptive field able to capture the marine context around the targets. YOLO-S is also extremely robust ranking first or second for almost every vehicle category. It is also the best performing CNN for the majority class "Car" and for the second largest class "Van". AP for majority class "Car" is instead degraded by as much as 10.9\% for YOLOv3 and 12.2\% for~\cite{Ju_2019}. This involves that wAP for YOLOv3 and~\cite{Ju_2019} is more than 10\% lower than YOLO-S. 

\begin{figure*}[!tb]
\centering
	\vspace{0cm}
	\includegraphics[width= \textwidth, keepaspectratio]{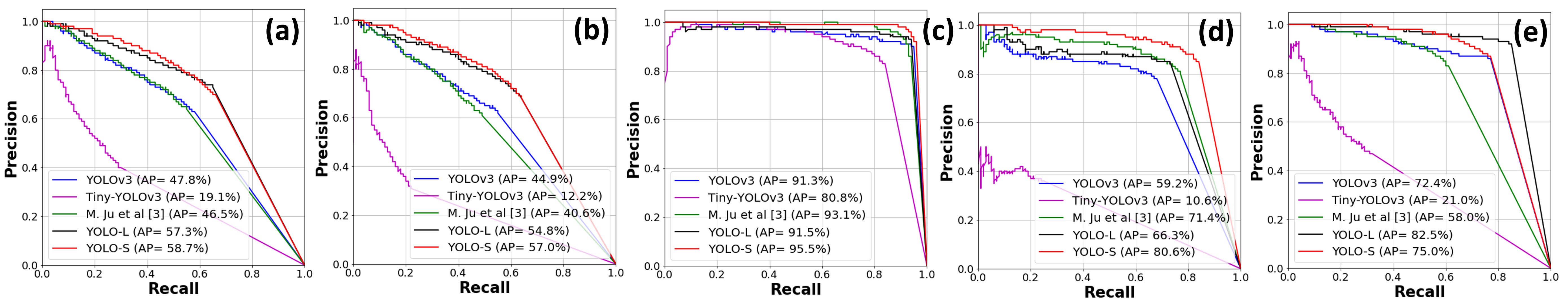}
	\vspace{-0.4cm}
	\caption{Precision-Recall curve of the proposed YOLO-L (black) and YOLO-S (red) and the baseline detectors YOLOv3 (blue), Tiny-YOLOv3 (magenta) and~\cite{Ju_2019} (green) for majority class "Car" for experiments (a) 1 and (b) 2 over AIRES dataset, (c) 1 and (d) 2 over VEDAI dataset and (e) for class "Person" in SARD dataset, respectively.} 
	\label{fig:prec_rec}
\end{figure*}

As expected, the broader the target size, the higher the COCO AP.
In particular, our CNNs play the lion’s share for all objects size, obtaining an AP ranging from 11.9\% for small objects to 29.3\% for large objects. YOLO-S outperforms the other detectors for medium-size targets. Our networks are also better when looking at micro-average accuracy. More in details, YOLO-L has the best trade-off between TPs and FPs, with a F1\textsubscript{ma} of 65.6\%. Nevertheless, YOLO-S is very competitive with a F1\textsubscript{ma} of 65.5\% and the best REC\textsubscript{ma} of 62.0\%. 
YOLOv3 is instead not adequately optimized for small target detection, as proven by the F1\textsubscript{ma} degraded by 7.3\% with respect to YOLO-L.  It is also confirmed that~\cite{Ju_2019} can almost catch up with performances of YOLOv3, although the  F1\textsubscript{ma} gap as compared to ours CNNs is  significant ($\approx$ 12\%). On the contrary, features extracted by Tiny-YOLOv3 are very poor, leading to a F1\textsubscript{ma} as low as 32.1\%, roughly one half of the proposed CNNs. 
Fig.~\ref{fig:prec_rec}a shows the precision-recall curves for the majority class "Car". It further highlights the competitiveness of the proposed networks and the similarity of performances of YOLOv3 and~\cite{Ju_2019}. Tiny-YOLOv3 is instead outperformed by the other networks by a wide margin. On the other hand, Tiny-YOLOv3 and~\cite{Ju_2019} are the fastest algorithms with 10.5 and 9.7 FPS, respectively, on the proposed hardware. However, despite of its simplicity, YOLO-S appears the best trade-off between accuracy, computing speed and memory required  with roughly 90\% of Tiny-YOLOv3 weights, and with a speed and a mAP improved by 37\% and 16\%, respectively, with respect to YOLOv3.

Tab.~\ref{tab:single_double_stage_training} illustrates results for experiment 2.
As it may be seen, when handling with a small dataset, features learned from a previous task impact on model performances and, in general, the more similar the source task to the target task, the higher the performances.
Specifically, using initial weights finetuned on domain related source images, such as DOTAv2 and VEDAI, can boost performances with respect to general features extracted in a more basic vision task such as COCO. 
The procedure is particularly beneficial for the micro-average metrics, despite of the usually different point of view of DOTAv2 and VEDAI images with respect to AIRES dataset. In particular, the relative recall (F1-score) improvement ranges from a few percent for YOLO-L, YOLO-S and YOLOv3 up to 10\% (8\%) for~\cite{Ju_2019} and 37\% (32\%) for Tiny-YOLOv3. The proposed networks outperform YOLOv3 also with a single-stage training and are remarkably better in terms of micro-average precision and recall. Also, Fig.~\ref{fig:prec_rec}b shows that for "Car" class YOLO-S slightly edges out YOLO-L, whereas YOLOv3 and~\cite{Ju_2019} have similar precision only at low recall rates. 
On the other hand, a single-stage fine-tuning does not result always in a performance drop of mAP and COCO metrics AP\textsubscript{S}, AP\textsubscript{M} and AP\textsubscript{L}, especially for YOLOv3 and YOLO-L. This is in part due to the sensitivity of such metrics on class imbalance, which is not managed in this work, since small absolute variations on detected instances of minority classes can lead to large fluctuations on the corresponding APs, thus affecting significantly the mean scores. This is confirmed by looking at wAP, which exhibits an absolute improvement for all networks by implementing a double-stage learning and ranging from almost 1\% for YOLO-L to 6\% for Tiny-YOLOv3. Hence, double-stage training increases model robustness and facilitates generalization beyond the training data, as shown by the increase in the overall accuracy. 
 
\begin{figure*}[!htb]
\small
\centering
	\vspace{0cm}
	\includegraphics[width= \textwidth, trim={1cm 0 0 0}, clip ,keepaspectratio]{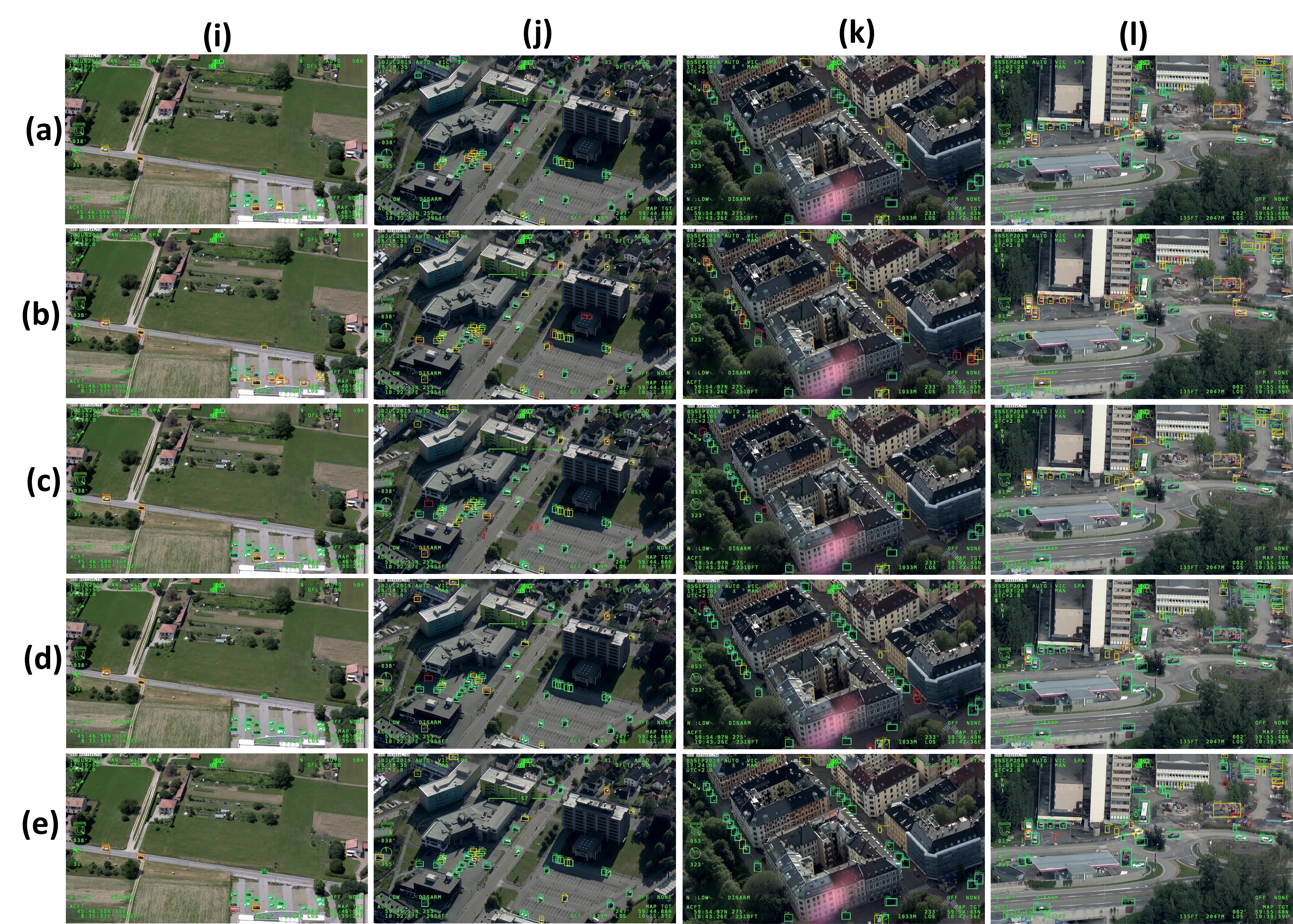}
	\vspace{0cm}
	\caption{Experiment 1 on AIRES dataset. Comparison of different CNNs: (a) YOLOv3, (b) Tiny-YOLOv3, (c) CNN by Ref.~\cite{Ju_2019}, (d) YOLO-L (ours), (e) YOLO-S (ours). The following background are considered: (i) rural (Italy), (j), (k), (l) urban (Norway).} 
	\label{fig:qualitative_our_exp_1}
\end{figure*}

\subsubsection{Qualitative evaluation}\label{Results_qualitative_our}
In Fig.~\ref{fig:qualitative_our_exp_1} we benchmark detection results of the five mentioned architectures for experiment 1 on four images characterized by different background, either rural or urban, and variable target size. Green (red) box denotes the true (false) positive TP (FP), whereas the ground truth box is depicted as a light blue box if detection of the related target is correct and as a yellow box if a false negative (FN) occurs.
It is evident that all networks outperform significantly Tiny-YOLOv3. Concerning rural scenario in Fig.~\ref{fig:qualitative_our_exp_1}i, both our CNNs outperform the other networks, with 2 (3) FNs for YOLO-L (YOLO-S), which have been correctly located but misclassified. Missed detections are 4 for~\cite{Ju_2019}, 5 for YOLOv3 and 11 for Tiny-YOLOv3, the latter having also a very poor target localization due to the lack of a fine-grained output scale. Similarly, FPs increase from 3 for YOLO-L and YOLOv3, 4 for YOLO-S and~\cite{Ju_2019}, up to 11 for Tiny-YOLOv3. 

Similar results are obtained for urban scenario in Fig.~\ref{fig:qualitative_our_exp_1}j: our networks perform as well as YOLOv3, with a number of missed detections and misclassified detections which increases significantly for the other networks. In particular, Tiny-YOLOv3 fails to detect the larger part of vehicles. 
Our networks have competitive performances also in scenarios with targets having larger size and weak contrast with ground (Fig.~\ref{fig:qualitative_our_exp_1}k), or moderate size variability (Fig.~\ref{fig:qualitative_our_exp_1}l).
Specifically, in Fig.~\ref{fig:qualitative_our_exp_1}k YOLO-S outperforms the other CNNs with only 4 missed targets and 1 false positive. YOLO-L,~\cite{Ju_2019} and YOLOv3 have also satisfactory performances, with 4, 5 and 6 missed detections, respectively. On the contrary, Tiny-YOLOv3 produces a large amount of false positives and false negatives.
In Fig.~\ref{fig:qualitative_our_exp_1}l the proposed CNNs play the lion's share once more and are able to detect some partially occluded targets or with poor texture information. 

\begin{table*}[!b]
\small
  \centering
    \begin{tabular}{cccccc}
    \hline
    \textbf{Metric} & \textbf{YOLOv3} & \textbf{Tiny-YOLOv3} & \textbf{M. Ju et al~\cite{Ju_2019}} & \textbf{YOLO-L} & \textbf{YOLO-S} \\
    \hline
    \textbf{Van [\%]} & \textcolor{green}{\textbf{62.6}} & 38.7  & 54.2  & 59.4 & \textbf{70.7}  \\
     \textbf{Truck [\%]} & 51.1 & 35.4 & 40.3   & \textcolor{green}{\textbf{60.3}} & \textbf{65.5} \\
     \textbf{Car [\%]} & 91.3  & 80.8 &  \textcolor{green}{\textbf{93.1}}  & 91.5  & \textbf{95.5} \\
     \textbf{Tractor [\%]} & 39.0  & 21.6 &  42.3 & \textbf{70.6}  & \textcolor{green}{\textbf{63.8}} \\
     \textbf{Other [\%]} & 30.7 & 33.6 & 25.0  & \textcolor{green}{\textbf{39.4}} & \textbf{47.9} \\
     \textbf{Plane [\%]} & 86.7 & \textcolor{green}{\textbf{89.9}}  & 81.1  & \textbf{98.7} & 75.0 \\
     \textbf{Boat [\%]} & \textcolor{green}{\textbf{65.2}} & 23.9  &  44.5 & 62.5 & \textbf{74.2} \\
    \hline
    \textbf{mAP [\%]} & 60.9  & 46.3    & 54.3    & \textcolor{green}{\textbf{68.9}} & \textbf{70.4} \\
    \hline
    \textbf{wAP [\%]} & 77.3 &	63.4 &	75.2 &	\textcolor{green}{\textbf{79.7}} &	\textbf{84.6}  \\
	\hline    
	 \textbf{AP\textsubscript{S} [\%]} & \textcolor{green}{\textbf{24.2}} & 9.4   & 16.4 & 22.8  & \textbf{30.8} \\
	\hline   
	 \textbf{AP\textsubscript{M} [\%]} & 35.4 & 25.0  &  31.4  & \textcolor{green}{\textbf{39.6}}  &  \textbf{41.8} \\
	\hline  
	 \textbf{AP\textsubscript{L} [\%]} & 30.3  &  23.4  & 23.9  & \textbf{39.0} &  \textcolor{green}{\textbf{31.9}} \\
	\hline  
    \textbf{REC\textsubscript{ma} [\%]} & 83.4  & 70.4  & 80.3  & \textcolor{green}{\textbf{84.9}} & \textbf{88.1} \\
	\hline    
    \textbf{PREC\textsubscript{ma} [\%]} & \textcolor{green}{\textbf{81.1}}  & 71.5  & 73.6  & \textbf{82.7} & 79.6 \\
	\hline   
    \textbf{F1\textsubscript{ma} [\%]} & 82.2  & 70.9  & 76.8  & \textbf{83.8} & \textcolor{green}{\textbf{83.6}} \\
	\hline  
	\textbf{FPS} & 6.4 & \textbf{13.0} & \textcolor{green}{\textbf{11.9}} &  6.3 & 9.7 \\
    \hline  
    \end{tabular}%
    \vspace{0.2cm}
    \caption{Experiment 1 on VEDAI dataset. Comparative results of YOLOv3~\cite{Redmon_2018}, Tiny-YOLOv3~\cite{YOLO}, CNN by Ref.~\cite{Ju_2019}, YOLO-L and YOLO-S in terms of mAP, wAP, AP\textsubscript{S}, AP\textsubscript{M}, AP\textsubscript{L}, REC\textsubscript{ma}, PREC\textsubscript{ma}, F1\textsubscript{ma} and FPS. The 1st (2nd) best score is highlighted in black (green) bold.}
  \label{tab:performance_VEDAI_dataset_exp_1}%
\end{table*}%

\begin{table*}[!tb]
\small
  \centering
    \begin{tabular}{cccccc}
    \hline
    \textbf{Metric} & \textbf{YOLOv3} & \textbf{Tiny-YOLOv3} & \textbf{M. Ju et al~\cite{Ju_2019}} & \textbf{YOLO-L} & \textbf{YOLO-S} \\
    \hline
    \textbf{Van [\%]} & 31.4 & 0.4 & 27.1 & \textcolor{green}{\textbf{38.2}} &  \textbf{44.5} \\
     \textbf{Truck [\%]} & 13.9 & 1.4 &  12.4 & \textcolor{green}{\textbf{30.1}} & \textbf{32.6} \\
     \textbf{Car [\%]} & 59.2 & 10.6 & \textcolor{green}{\textbf{71.4}}  & 66.3  & \textbf{80.6} \\
     \textbf{Tractor [\%]} & 4.3 & 0.5 &  \textcolor{green}{\textbf{4.5}}  & 3.5  & \textbf{27.3} \\
     \textbf{Other [\%]} & 15.2 & 0.7 & 9.6 & \textcolor{green}{\textbf{16.0}} & \textbf{19.9} \\
     \textbf{Plane [\%]} & \textbf{96.1} & 26.4  & 57.3 & 62.9 & \textcolor{green}{\textbf{74.2}} \\
     \textbf{Boat [\%]} & \textcolor{green}{\textbf{7.9}} & 1.0  &  3.9 & 4.0 & \textbf{25.9} \\
    \hline
    \textbf{mAP [\%]} & \textcolor{green}{\textbf{32.6}} &  5.8 &  26.6 & 31.6 & \textbf{43.6} \\
     \hline
    \textbf{wAP [\%]} & 44.8 & 7.2 & 50.8 &	50.9 &	63.5 \\
	\hline 
	\textbf{AP\textsubscript{S} [\%]} & 6.7 & 0.9  & 6.1  & \textcolor{green}{\textbf{8.7}}  & \textbf{9.4} \\
	\hline   
	 \textbf{AP\textsubscript{M} [\%]} & 10.0 &  1.4 & 12.1 & \textcolor{green}{\textbf{13.5}}  & \textbf{22.2} \\
	\hline  
	 \textbf{AP\textsubscript{L} [\%]} & \textcolor{green}{\textbf{23.6}} &  6.9  & 13.9  & \textbf{30.6} &  22.8 \\
	\hline     
    \textbf{REC\textsubscript{ma} [\%]} & 52.8 & 16.7 & 57.3 & \textcolor{green}{\textbf{57.5}} & \textbf{68.8} \\
	\hline    
    \textbf{PREC\textsubscript{ma} [\%]} & 71.6  & 33.5 & 74.2 & \textbf{78.9} & \textcolor{green}{\textbf{78.0}} \\
	\hline   
    \textbf{F1\textsubscript{ma} [\%]} & 60.8 & 22.3 & 64.6 & \textcolor{green}{\textbf{66.6}} & \textbf{73.1} \\
	\hline   
	\textbf{FPS} & 18.5 & \textbf{31.6} & \textcolor{green}{\textbf{29.8}} &  18.3 & 25.6 \\
    \hline   
    \end{tabular}%
    \vspace{0.2cm}
    \caption{Experiment 2 on VEDAI dataset. Comparative results of YOLOv3~\cite{Redmon_2018}, Tiny-YOLOv3~\cite{YOLO}, CNN by Ref.~\cite{Ju_2019}, YOLO-L and YOLO-S in terms of mAP, wAP, AP\textsubscript{S}, AP\textsubscript{M}, AP\textsubscript{L}, REC\textsubscript{ma}, PREC\textsubscript{ma},  F1\textsubscript{ma} and FPS. The 1st (2nd) best score is highlighted in black (green) bold.}
  \label{tab:performance_VEDAI_dataset_exp_2}%
\end{table*}%

\subsection{Experiments on VEDAI dataset}
In Experiment 1 a moving window of size 537$\times$537 pixel slides across each input VEDAI image with a sliding step equal to 487 pixels, whereas a "single-shot" inference on the whole input image has been employed for Experiment 2. Besides, a double-stage (single-stage) training has been implemented in Experiment 1 (2).
 
\subsubsection{Quantitative comparison with State of the Art detectors}\label{Results_quantitative}

The experimental results for each algorithm are illustrated in Tab.~\ref{tab:performance_VEDAI_dataset_exp_1}.
All architectures outperform largely Tiny-YOLOv3, even if in this case the latter can achieve a decent trade-off between precision and recall (Fig.~\ref{fig:prec_rec}c). 
Our CNNs have the most competitive results once more with mAP values of 70.4\% and 68.9\% for YOLO-S and YOLO-L, respectively. YOLO-S has a better average precision for the three classes having broader statistics, i.e. "Car", "Van" and "Truck", and for small and medium-size objects. 
Further, YOLO-S outperforms the other CNNs in term of micro recall, whereas YOLO-L exhibits the highest balancing between precision and recall. Specifically, F1\textsubscript{ma} is 83.8\% for YOLO-L, 83.6\% for YOLO-S and 82.2\% for YOLOv3. The other detectors are instead lagging behind, with a degradation of 7\% for~\cite{Ju_2019} and almost 13\% for Tiny-YOLOv3. Nevertheless, CNN by Ref.~\cite{Ju_2019} is quite close to YOLOv3, with a mAP (wAP) drop equal to 6.6\% (2.1\%). It can also achieve a better precision than YOLOv3 on class "Car" while only sacrificing minimal recall (Fig.~\ref{fig:prec_rec}c). 
YOLO-S is striking the best balance between accuracy and speed: it is up to 16\% more accurate and 52\% faster than YOLOv3, resulting only about 25\% (18\%) slower than Tiny-YOLOv3 (\cite{Ju_2019}). 

When comparing average performances with those in Tab.~\ref{tab:performance_our_dataset}, they are significantly better on VEDAI dataset even with a smaller statistics available. Indeed all VEDAI images have been captured at nadir and at the same distance to the ground, unlike AIRES dataset where the point of view and acquisition height change significantly depending on camera tilt angle and  helicopter altitude and making consequently more challenging the detection. In addition, moving window size is smaller for VEDAI images, which facilitates target localization. 

In Tab.~\ref{tab:performance_VEDAI_dataset_exp_2} we show results for Experiment 2. 
In this case, after image resize to 416$\times$416, the effective median target dimension spans from 13$\times$13 for "Car" up to 37$\times$37 for "Plane", much smaller than receptive fields of YOLOv3 (Tab.~\ref{tab:cnn_properties}) and making more challenging feature extraction for all CNNs due to the poor number of meaningful pixels after many convolution operations. 
As a consequence, performances degrade for each algorithm with a severe mAP drop by a factor 8 for Tiny-YOLOv3 and limited ($\approx$ 1.6) for YOLO-S. 
In particular, YOLO-S exhibits still satisfactory results and outperforms all the other networks, with a mAP of 43.6\% and a F1\textsubscript{ma} of 73.1\%. 
It shows also the best balancing between REC\textsubscript{ma} (68.8\%) and PREC\textsubscript{ma} (78.0\%), as also highlighted by Fig.~\ref{fig:prec_rec}d for the majority class, which indicates that YOLO-S discriminates well the artifacts from real vehicles instances. 
Compared to YOLOv3, the mAP, micro F1-score, recall and precision have a relative improvement of 33.7\%, 20.2\%, 30.3\% and 8.9\%, respectively. YOLO-S is also the best detector for almost all vehicle classes and object size, with a true positive detection rate more than 10\% higher than the second best performing network YOLO-L. As in experiment 1, Tiny-YOLOv3 is significantly outperformed by all the other algorithms, including CNN by Ref.~\cite{Ju_2019} which in turn performs better than YOLOv3 for all the considered micro-average metrics and also for wAP. Indeed, on majority class, network by Ref.~\cite{Ju_2019} can achieve higher precision for comparable recall (Fig.~\ref{fig:prec_rec}d). 

Tiny-YOLOv3 is the fastest network processing almost 31.6 FPS. Nevertheless, YOLO-S can process as many FPS as 25.6, resulting roughly 38\% faster than YOLOV3 and only the 19\% (14\%) slower than Tiny-YOLOv3 (\cite{Ju_2019}). 
This confirms that YOLO-S is the best trade-off also on VEDAI dataset.

\begin{figure*}[!tb]
\small
\centering
	\vspace{0cm}
	\includegraphics[width= 0.98\textwidth, trim={1cm 0 0 0}, clip ,keepaspectratio]{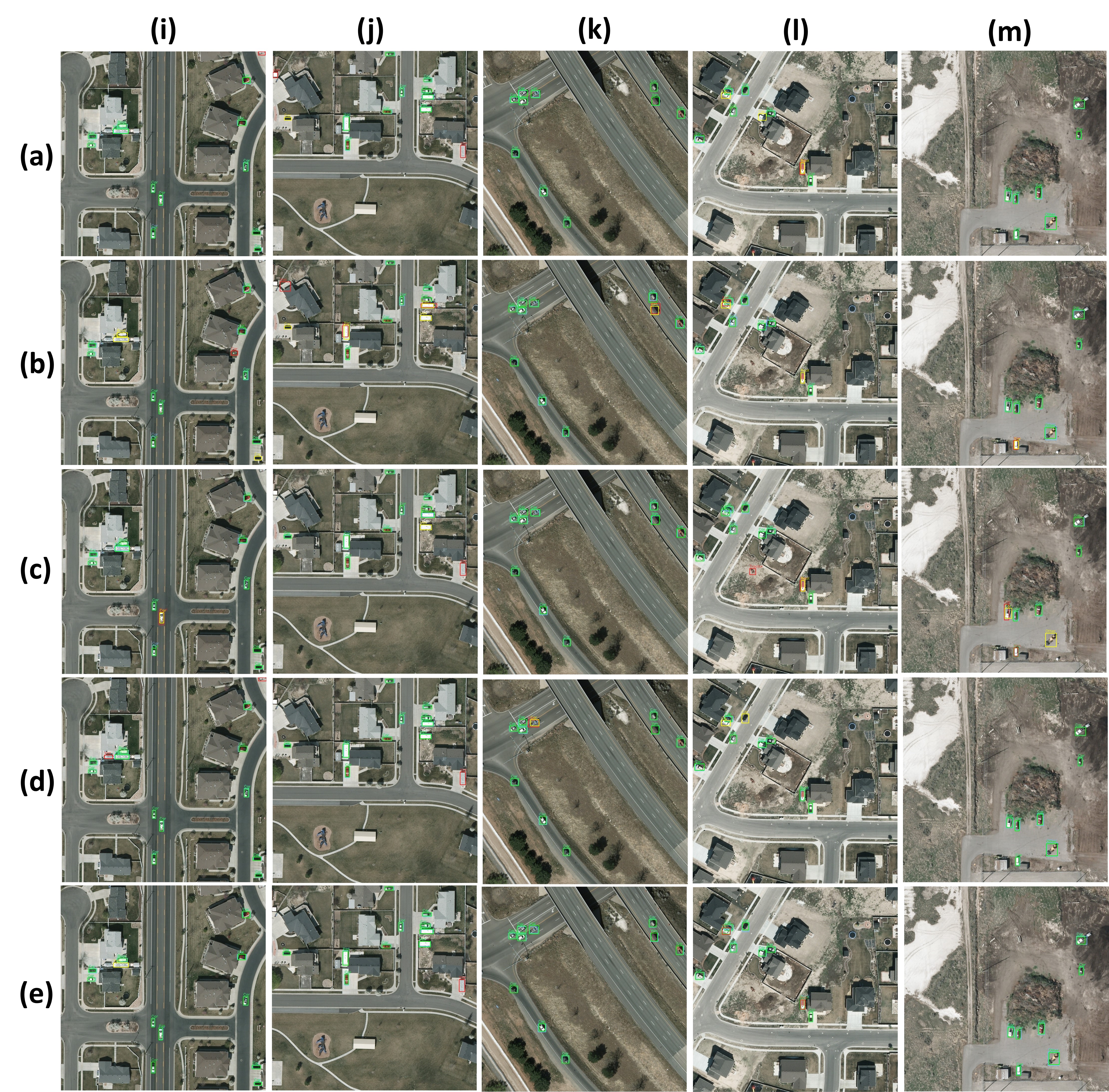}
	\vspace{0cm}
	\caption{Experiment 1 on VEDAI dataset. Comparison of different CNNs: (a) YOLOv3, (b) Tiny-YOLOv3, (c) CNN by Ref.~\cite{Ju_2019}, (d) YOLO-L (ours), (e) YOLO-S (ours).} 
	\label{fig:qualitative_vedai_exp_1}
\end{figure*}

\begin{figure*}[!tb]
\small
\centering
	\vspace{0cm}
	\includegraphics[width= 0.98\textwidth, trim={1cm 0 0 0}, clip ,keepaspectratio]{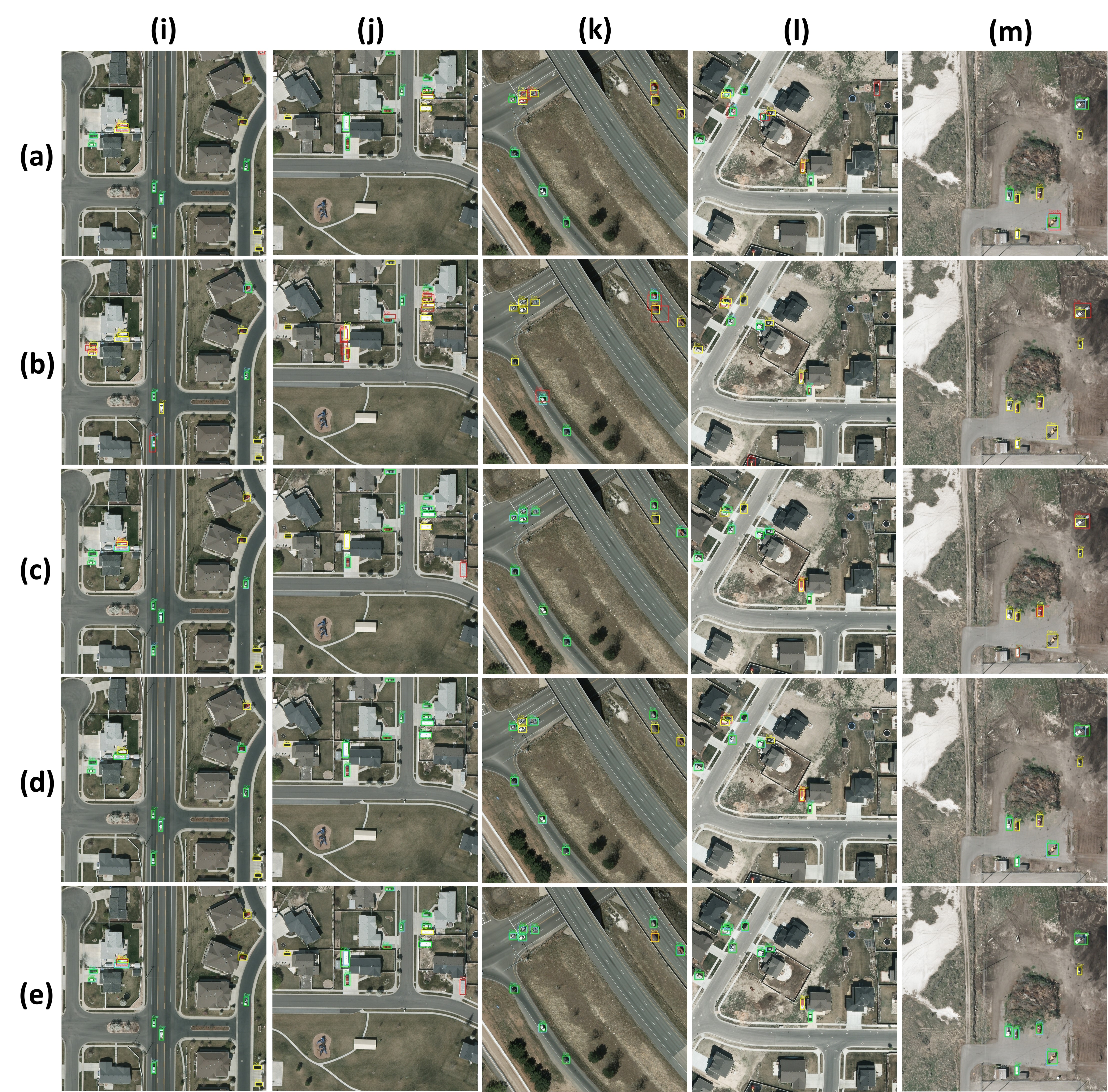}
	\vspace{0cm}
	\caption{Experiment 2 on VEDAI dataset. Comparison of different CNNs: (a) YOLOv3, (b) Tiny-YOLOv3, (c) CNN by Ref.~\cite{Ju_2019}, (d) YOLO-L (ours), (e) YOLO-S (ours).} 
	\label{fig:qualitative_vedai_exp_2}
\end{figure*}

\subsubsection{Qualitative evaluation}\label{Results_qualitative_vedai}
Fig.~\ref{fig:qualitative_vedai_exp_1} shows the detection results for experiment 1 on five different test images. 
It may be seen that YOLOv3, YOLO-L, YOLO-S and \cite{Ju_2019} have all remarkable performances and outperform Tiny-YOLOv3.
In particular, YOLO-S is the best performing network with only one missed detection in Fig.~\ref{fig:qualitative_vedai_exp_1}i out of five images. 
YOLOv3 does not produce any FNs in Fig.~\ref{fig:qualitative_vedai_exp_1}i, k and m. Similarly, YOLO-L detects every target in Fig.~\ref{fig:qualitative_vedai_exp_1}i, j and m, and misses only a few vehicles in (k) and (l). Network by Ref.~\cite{Ju_2019} makes also very few false negative and false positive samples.

In Fig.~\ref{fig:qualitative_vedai_exp_2} comparative results are shown instead for experiment 2 considering the same input images of Fig.~\ref{fig:qualitative_vedai_exp_1}.
A generalized degradation occurs for every network, as already remarked, and appears more severe for Tiny-YOLOv3, YOLOv3,~\cite{Ju_2019} and YOLO-L, in this order. 
YOLO-S achieves better performances for three out of five scenarios, with YOLO-L resulting the best performing method for the remaining two. In particular, YOLO-S is robust against vehicles with orientation as in Fig.~\ref{fig:qualitative_vedai_exp_2}k, like~\cite{Ju_2019}. 
It misses also just one detection in Fig.~\ref{fig:qualitative_vedai_exp_2}l and m, and yields a few FNs in Fig.~\ref{fig:qualitative_vedai_exp_2}i and j. Moreover, the number of misclassified samples is negligible. 
CNN by Ref.~\cite{Ju_2019} performs better or equal to YOLOv3 in four out of five scenarios, whereas it fails in Fig.~\ref{fig:qualitative_vedai_exp_2}m. 
Further, all networks outperform Tiny-YOLOv3 by a wide margin, the latter resulting not adequate for any of the challenging scenarios discussed in Fig.~\ref{fig:qualitative_vedai_exp_2}.

%

\begin{table*}[!t]
\small
  \centering
    \begin{tabular}{cccccc|c}
    \hline
    \textbf{Metric} & \textbf{YOLOv3} & \textbf{Tiny-YOLOv3} & \textbf{M. Ju et al~\cite{Ju_2019}} & \textbf{YOLO-L} & \textbf{YOLO-S} & \textbf{YOLOv4~\cite{Sambolek_2021}}\\
    \hline
    \textbf{AP [\%]} & 72.4 &  21.0 &  58.0 & \textbf{82.5} & \textcolor{green}{\textbf{75.0}} & 88.0 \\
	\hline 
	\textbf{AP\textsubscript{S} [\%]} & \textcolor{green}{\textbf{13.4}} & 1.6  & 4.4  & \textbf{18.9 } & 11.2 & 26.0 \\
	\hline   
	 \textbf{AP\textsubscript{M} [\%]} & 32.9 &  5.6 & 28.2 & \textcolor{green}{\textbf{41.5}}  & \textbf{42.7} & 59.0  \\
	\hline  
	 \textbf{AP\textsubscript{L} [\%]} & 45.3 &  23.7  & 47.0  & \textcolor{green}{\textbf{55.6}} &  \textbf{61.2} & 75.0 \\
	\hline     
    \textbf{REC\textsubscript{ma} [\%]} & \textcolor{green}{\textbf{77.4}} & 31.6 & 61.1 & \textbf{84.7} & \textcolor{green}{\textbf{77.4}} & 89.8 \\
	\hline    
    \textbf{PREC\textsubscript{ma} [\%]} & 85.6  & 46.9 & 82.9 & \textbf{92.2} & \textcolor{green}{\textbf{87.1}} & 92.7 \\
	\hline   
    \textbf{F1\textsubscript{ma} [\%]} & 81.3 & 37.7 & 70.4 & \textbf{88.3} & \textcolor{green}{\textbf{82.0}} & 91.2 \\
	\hline   
	\textbf{FPS} & 17.3 & \textbf{27.8} & \textcolor{green}{\textbf{26.3}} &  16.9 & 22.8 & NA\\
    \hline   
    \end{tabular}%
    \vspace{0.2cm}
    \caption{Experiment on SARD dataset. Comparative results of YOLOv3~\cite{Redmon_2018}, Tiny-YOLOv3~\cite{YOLO}, CNN by Ref.~\cite{Ju_2019}, YOLO-L and YOLO-S in terms of AP, AP\textsubscript{S}, AP\textsubscript{M}, AP\textsubscript{L}, REC\textsubscript{ma}, PREC\textsubscript{ma},  F1\textsubscript{ma} and FPS. The 1st (2nd) best score is highlighted in black (green) bold. Best results from Ref.~\cite{Sambolek_2021} for YOLOv4 detector are also shown on the right.}
  \label{tab:performance_SARD_dataset_exp_1}%
\end{table*}%

\begin{figure*}[!b]
\small
\centering
	\vspace{0cm}
	\includegraphics[width= 0.98\textwidth, trim={1cm 0 0 0}, clip ,keepaspectratio]{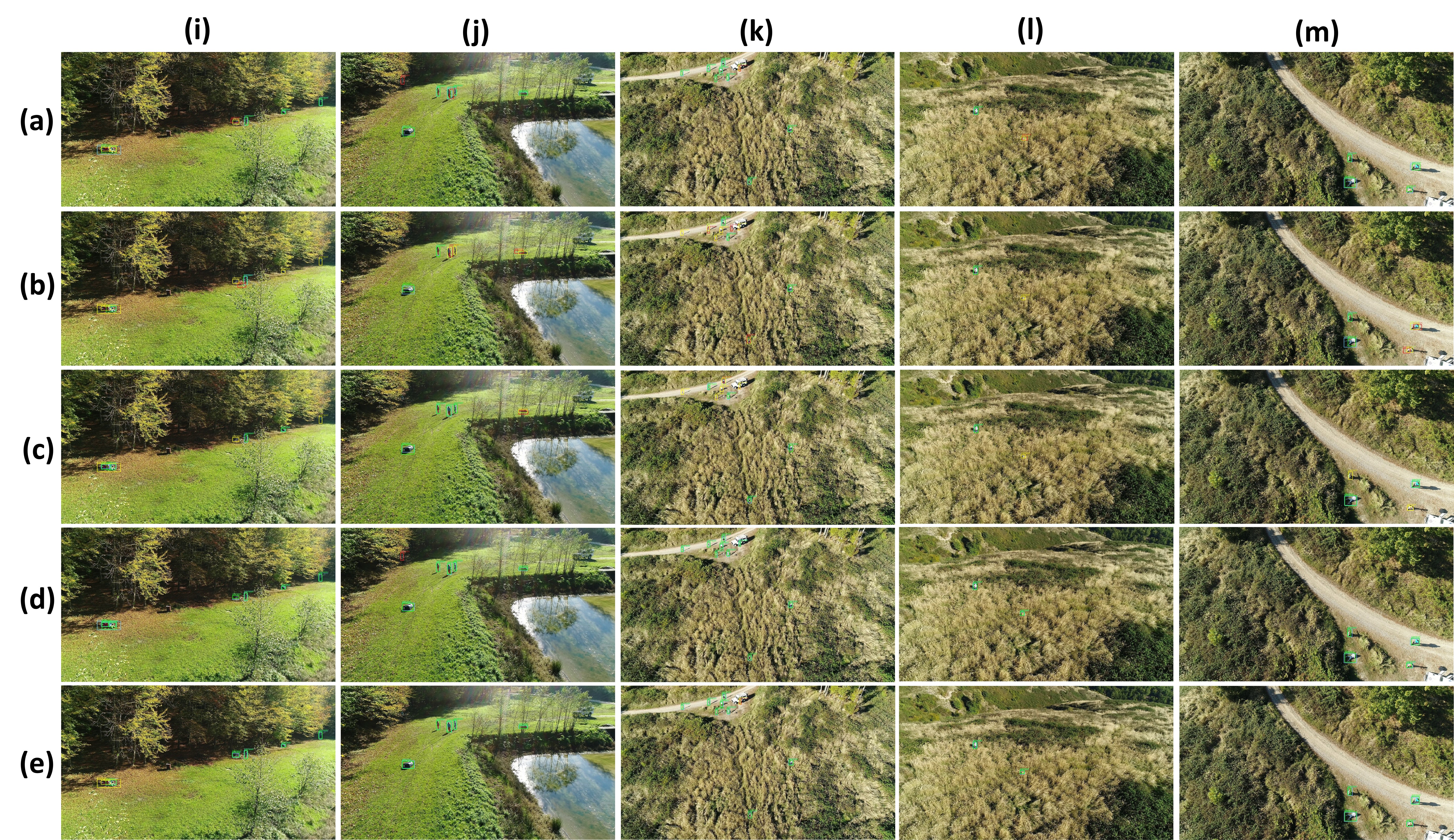}
	\vspace{0cm}
	\caption{Experiment on SARD dataset. Comparison of different CNNs: (a) YOLOv3, (b) Tiny-YOLOv3, (c) CNN by~\cite{Ju_2019}, (d) YOLO-L (ours), (e) YOLO-S (ours).} 
	\label{fig:qualitative_SARD}
\end{figure*}

\subsection{Experiment on SARD dataset}
The full-sized input image 1920$\times$1080 has been automatically resized to the network size 416$\times$416 pixel. Further, pre-trained weights on COCO have been employed for all architectures.
 
\subsubsection{Quantitative comparison with State of the Art detectors}\label{Results_quantitative}
Tab.~\ref{tab:performance_SARD_dataset_exp_1} shows the accuracy and speed performances for all the aforementioned networks. 
Despite of the different context, the proposed networks outperform once again the others for nearly all the detection metrics. In this case YOLO-L has the highest AP of more than 82\% and the best trade-off between precision and recall (Fig.~\ref{fig:prec_rec}e). Such values @0.5 are also quite competitive with the YOLOv4 detector~\cite{Sambolek_2021}, that shows the same precision and a slightly better F1-score, whereas for higher IoU thresholds the gap is larger. It is worth noticing however that YOLOv4 has been trained on image resolution 512$\times$512 and tested on 416$\times$416, which may increase a bit performances. 

Nevertheless, YOLO-S demonstrates again an outstanding detection ability with an AP increase of 2.6\% with respect to YOLOv3 and a higher positive predictive value for comparable true positive rate (Fig.~\ref{fig:prec_rec}e). In addition, the highest AP\textsubscript{M} for the most populated category of medium-size objects demonstrates a tighter matching between predicted bounding boxes and ground truth boxes for more stringent IoU thresholds. 
Tiny-YOLOV3 is instead by far the worst network with an AP suppressed by a factor more than 2.7 (3.5) with respect to~\cite{Ju_2019} (YOLO-S), hence resulting not adequate for SAR operations. 
The detection speed improvement of YOLO-S is also noticeable, resulting 31.8\% faster than YOLOv3.

\subsubsection{Qualitative evaluation}\label{Results_qualitative_vedai}
In Fig.~\ref{fig:qualitative_SARD} we qualitatively inspect the behaviour of the different networks on five test images. 
YOLO-L exhibits nearly perfect detection performances of human targets located at variable depth, having different positions such as standing, lying or crouched, as well as with a few distinctive features and poor contrast since hidden in high grass (Fig.~\ref{fig:qualitative_SARD}k, l), surrounded by a shady forest (Fig.~\ref{fig:qualitative_SARD}i) or wearing light-coloured clothes into a similar coloured dirt path (Fig.~\ref{fig:qualitative_SARD}m). Netherless, also YOLO-S enjoys very satisfactory results with just one missed detection in Fig.~\ref{fig:qualitative_SARD}k, which has been correctly identified but poorly located, and two in Fig.~\ref{fig:qualitative_SARD}i. 
On the other hand, YOLOv3 tends sometimes to miss or misdetect some targets. This is because of vague features extracted after many forward convolutional layers leading to FNs (Fig.~\ref{fig:qualitative_SARD}i, k, l), and the coarse granularity of two output scales which may yield FPs due to poor localization of real humans (Fig.~\ref{fig:qualitative_SARD}i, j, k, l). 
The problem is of course much more severe for Tiny-YOLOv3, which is unable to detect the larger part of persons due to the absence of the YOLOv3 finest grained scale 52$\times$52.

\begin{figure*}[!tb]
\small
\centering
	\vspace{0cm}
	\includegraphics[width= \textwidth, trim={0 0 0 0}, clip,keepaspectratio]{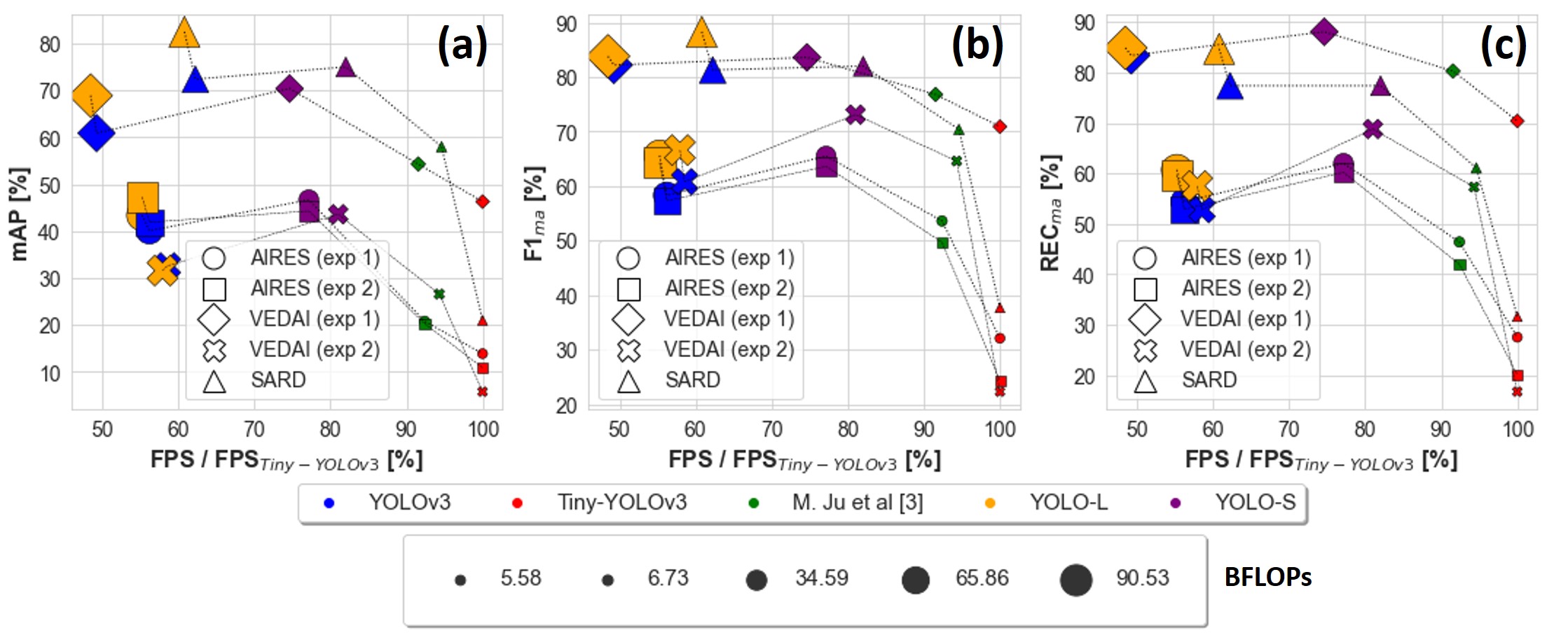}
	\vspace{-0.5cm}
	\caption{Performance summary of the different networks (blue: YOLOv3~\cite{Redmon_2018}, red: Tiny-YOLOv3~\cite{YOLO}, green: CNN by Ref.~\cite{Ju_2019}, orange: YOLO-L, violet: YOLO-S) on AIRES, VEDAI and SARD datasets: (a) mAP, (b) F1\textsubscript{ma} and (c) REC\textsubscript{ma} as a function of FPS. Same marker type corresponds to same experiment and dataset: circle (square) for experiment 1 (2) on AIRES dataset, diamonds ("x") for experiment 1 (2) on VEDAI dataset and triangle for experiment on SARD dataset. The marker size is proportional to the BFLOPs required by the network. For a better readability the FPS is normalized with respect to the Tiny-YOLOv3's processing speed.} 
	\label{fig:results_summary}
\end{figure*}

\section{Conclusion}\label{sec13}

In this work we have proposed two novel YOLO-like networks designed specifically for small target detection from Aerial Imagery: YOLO-L and YOLO-S.
YOLO-L is similar to YOLOv3, but replaces the coarse-grained output scale 13$\times$13 with the finer-graned 104$\times$104 and occupies almost one third the volume: because of a latency time close to YOLOv3, it results suitable only for offline data processing and in this work has been exploited for benchmarking purpose. 

YOLO-S is instead a fast, lightweight (smaller than Tiny-YOLOv3) and accurate network with a single output scale 52$\times$52, employing residual connection blocks to preserve from vanishing gradient issue, as well as skip connection via lateral concatenation and reshape - passthrough layer to encourage feature reuse across network and improve consequently localization performances.
In addition, we have presented an in-depth quantitative and qualitative benchmarking of our networks with three baseline detectors, namely YOLOv3, Tiny-YOLOv3 and the architecture designed in Ref.~\cite{Ju_2019}, on our novel AIRES dataset of 1275 FHD aerial images and including more than 15 thousands annotated objects, as well as on VEDAI and SARD datasets.

Experiments have been conducted by implementing a detection pipeline based on a sliding window approach and a NMS post-processing to handle full-sized images and by means of a single-shot inference on the whole VEDAI and SARD images.
The main results are summarized in Fig.~\ref{fig:results_summary} for metrics mAP, micro-average recall and F1-score and can be listed as follows: \\
(i) By means of a sliding window method, as in experiment 1 on AIRES and VEDAI datasets, YOLO-S is almost 37\% (52\%) faster and has a relative mAP improvement of almost 16.5\% (15.6\%) with respect to YOLOv3 on AIRES (VEDAI) dataset (Fig.~\ref{fig:results_summary}a). \\
(ii) YOLO-S is only about 23\% (25\%) slower than Tiny-YOLOv3, but has a relative F1\textsubscript{ma}@0.5 increase by 104.0\% (17.9\%) on AIRES (VEDAI) dataset (Fig.~\ref{fig:results_summary}b). \\
(iii) The use of pre-trained weights on a source task more similar to the target domain corresponding to AIRES dataset, such as the publicly-available DOTAv2 and VEDAI (experiment 1), lead always to 
a relative increase in the overall micro-average metrics @0.5 with respect to more basic preliminary features extracted on COCO dataset (experiment 2) from a few percent up to more than 30\% for Tiny-YOLOv3 (Fig.~\ref{fig:results_summary}b-c). \\
(iv) By applying inference on the full-sized VEDAI images with preliminary features learned on the basic COCO task (experiment 2), a F1\textsubscript{ma}@0.5 performance drop occurs for all networks with respect to point~(i), resulting very severe for Tiny-YOLOv3 (-48.3\%), significant for YOLOv3 and YOLO-S ($\approx$ -20\%) and moderate for~\cite{Ju_2019} and YOLO-S, resulting -10.5\% for the latter (Fig.~\ref{fig:results_summary}b). \\
(v) YOLO-S outperforms all the other detectors for nearly all the considered metrics (Fig.~\ref{fig:results_summary}a-b-c). \\
(vi) YOLO-S is able to process almost 25.6 FPS on the proposed hardware, resulting 38\% faster than YOLOv3 and just 19\% (14\%) slower than Tiny-YOLOv3 (\cite{Ju_2019}) (Fig.~\ref{fig:results_summary}a-b-c). \\
(vii) Inference on full-sized SARD dataset demonstrates that YOLO-S is also able to generalize well to a different context such as the search for injured people or people in a life threatening situation, with a slightly better detection accuracy than YOLOv3 and with a 32\% improvement in the processing speed (Fig.~\ref{fig:results_summary}a-b). Among the discussed baseline detectors, only YOLO-L has a better average precision, but with a significantly slower detection speed and almost 2.6 times larger BFLOPs (Fig.~\ref{fig:results_summary}a). \\
(viii) Regardless of experiments and datasets, YOLO-S is just from 15\% to 25\% slower than Tiny-YOLOv3, but even more accurate than YOLOv3 and YOLO-L. \\ 
(ix) YOLO-S has about one half BFLOPs of YOLOv3, quite close to SlimYOLOv3-SPP3-50~\cite{Zhang_2019}. As a consequence, as far as power consumption is concerned like for example for low-power real-time UAV applications, it is an interesting and less voluminous alternative to the by far less precise Tiny-YOLOv3. 

In summary, despite of its simplicity, results demonstrate that YOLO-S can really meet both accuracy and real-time detection requirements and is consequently a very promising candidate for integration on low-power GPU-less systems.  
Next steps include the deployment of YOLO-S and investigation of data quantization techniques~\cite{Vestias_2019} to reduce further the computational complexity and make the network even smaller and faster.


\section*{Author Contributions}{A.B.: conceptualization, methodology, software, validation, experiments, visualization, writing, review and editing. The author has read and agreed to the published version of the manuscript.}



\section*{Acknowledgment}
This research did not receive any specific grant from funding agencies in the public, commercial, or
not-for-profit sectors. 
This work is a follow-on of the results discussed in the EDA workshop on “Emerging \& disruptive technologies in optronics for defence applications” held in Brussels on 18-19/10/2021. 
The author acknowledges the R\&D Department of the company FlySight S.r.l. (\url{https://www.flysight.it/}) for the support received in the identification of the operative scenario, for data procurement and the help in the image annotation process.

\end{document}